\documentclass[10pt,twocolumn,letterpaper]{article}
\usepackage[pagenumbers]{iccv}

\definecolor{iccvblue}{rgb}{0.21,0.49,0.74}
\usepackage[pagebackref,breaklinks,colorlinks,allcolors=iccvblue]{hyperref}
\usepackage{couriers}
\usepackage{multirow}
\usepackage{threeparttable}
\usepackage{amssymb}
\usepackage{pifont}
\usepackage{xcolor}

\def\paperID{}
\def\confName{ICCV}
\def\confYear{2025}

\newcommand{\model}{\textsc{CaRe}}
\newcommand{\benchmark}{\textsc{CaReBench}}
\newcommand{\cmark}{\ding{51}}
\newcommand{\xmark}{\ding{55}}
\usepackage{soul}
\usepackage{tcolorbox}
\usepackage{color}
\usepackage{marvosym}
\tcbuselibrary{breakable}
\newtcolorbox{textbox}[1]{
    sharp corners,
    boxsep=0mm,
    toptitle=2mm,
    lefttitle=0mm,
    colframe=black!3,
    colback=black!3,
    title={\rule[-2pt]{4.5pt}{10pt}\hspace*{1.5mm}#1},
    breakable=true,
    fonttitle=\bfseries\itshape\sffamily,
    coltitle=black,
}

\title{\benchmark: A Fine-Grained Benchmark for Video Captioning and Retrieval}

\author{
    Yifan Xu$^{1}$, 
    Xinhao Li$^{1,2}$, 
    Yichun Yang$^1$, 
    Desen Meng$^1$, 
    Rui Huang$^1$, 
    Limin Wang$^{1,2,}$\textsuperscript{\Letter} \\[3pt]
    \small$^{1}$State Key Laboratory for Novel Software Technology, Nanjing University \hspace{0.5cm} \\
    \small $^{2}$Shanghai AI Laboratory \hspace{0.5cm} \\[3pt]
    {\small\url{https://carebench.github.io}}
}

\begin{document}
\makeatletter
\let\@oldmaketitle\@maketitle
\renewcommand{\@maketitle}{\@oldmaketitle
    \centering
    \includegraphics[width=\linewidth]{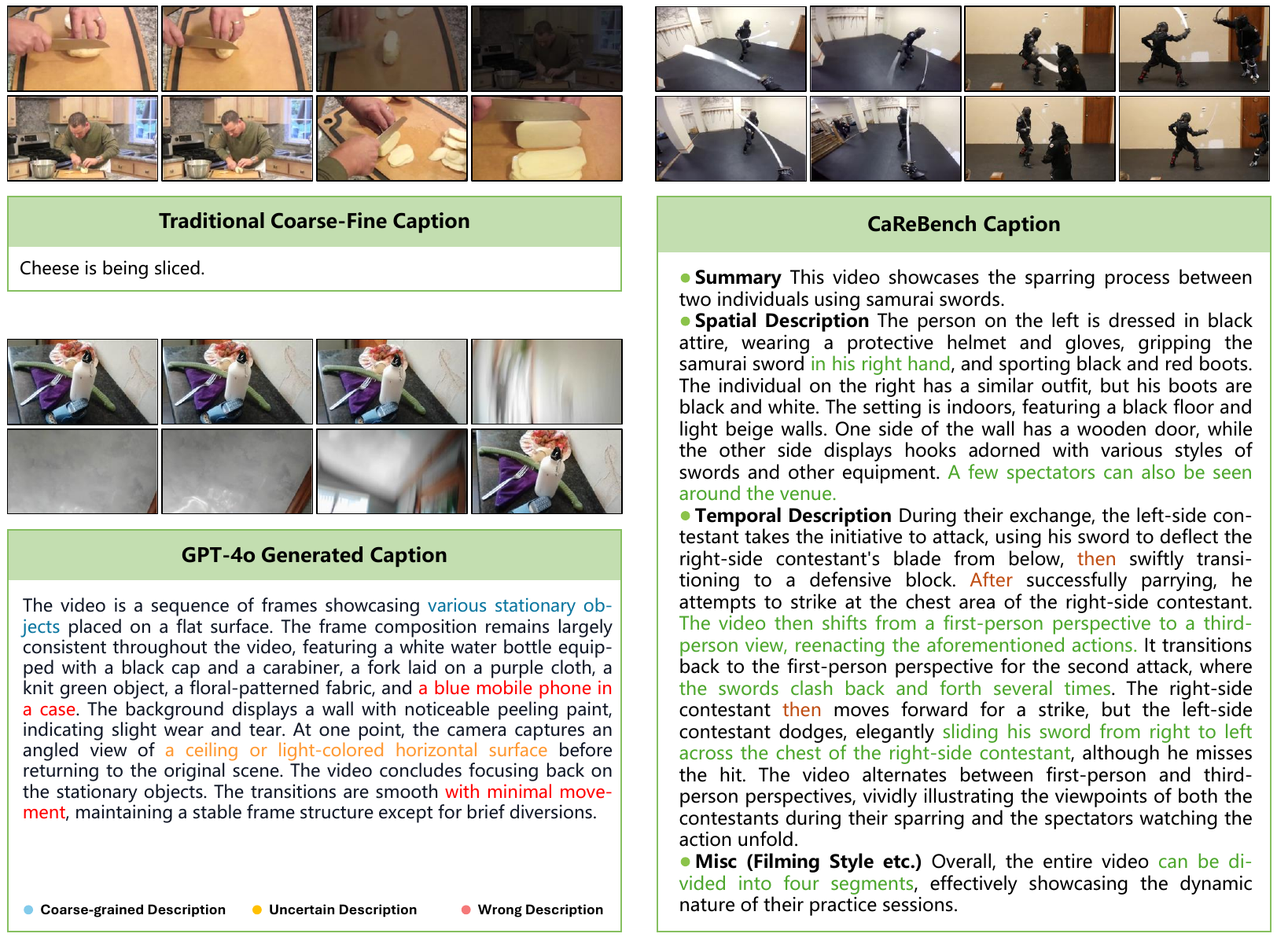}
    \captionof{figure}{\textbf{Comparision of captions between MSR-VTT~\cite{MSRVTT}, GPT-4o generated data~\cite{sharegpt4o} and \benchmark{}}. The caption in the upper left corner is from MSR-VTT~\cite{MSRVTT}. It only contains short-text coarse descriptions. The annotation located in the lower left corner is generated by GPT-4o sourced from ShareGPT-4o~\cite{sharegpt4o}. It has some \textcolor[RGB]{147,201,231}{coarse-grained}, \textcolor[RGB]{246,194,66}{uncertain} and \textcolor[RGB]{229,128,125}{wrong} descriptions. The fine-grained caption on the right is selected from \benchmark{} and is created by our human annotators following the pipeline. The \textcolor[RGB]{101,165,66}{green} sentences are fine-grained descriptions and the \textcolor[RGB]{179,86,40}{brown} words show the temporal sequences in the video.}
    \label{fig:caption-compare}
   \bigskip}
\makeatother
\maketitle
{
\renewcommand{\thefootnote}%
{\fnsymbol{footnote}}
\footnotetext[0]{\Letter{} Corresponding author.} 
}
\clearpage
\begin{abstract}
Video understanding, including video captioning and retrieval, is still a great challenge for video-language models (VLMs). The existing video retrieval and caption benchmarks only include short descriptions, limits their ability of detailed video understanding evaluation. To address this problem, we present \benchmark, a testing benchmark for fine-grained video \textbf{Ca}ptioning and \textbf{Re}trieval with 1,000 high-quality pairs of videos and human-annotated detailed captions. Uniquely, it provides manually separated spatial annotations and temporal annotations for each video. Based on this design, we introduce two evaluation metrics, ReBias and CapST, specifically tailored for video retrieval and video captioning tasks, respectively. These metrics enable a comprehensive investigation into the spatial and temporal biases inherent in VLMs. In addition, to handle both video retrieval and video captioning tasks in a unified framework, we develop a simple baseline based on a Multimodal Language Model (MLLM). By implementing a two-stage Supervised Fine-Tuning (SFT), we fully unlock the potential of MLLM, enabling it not only to generate detailed video descriptions but also to extract video features. Surprisingly, experimental results demonstrate that, compared to the CLIP-based models designed for retrieval and the popular MLLMs skilled in video captioning, our baseline shows competitive performance in both fine-grained video retrieval and video detailed captioning. 
\end{abstract}
    
\section{Introduction}
\label{sec:intro}

\begin{table*}
\centering
\resizebox{\textwidth}{!}{
\begin{tabular}{lcccccccc}
\toprule
\textbf{Benchmark} & \multicolumn{1}{l}{\textbf{\# Sample}} & \multicolumn{1}{c}{\textbf{Avg. Len.}} & \multicolumn{1}{c}{\textbf{Avg. Words}} & \multicolumn{1}{c}{\textbf{Annotator}} & \multicolumn{1}{c}{\textbf{Diverse Anno.}} & \textbf{Static Focus} & \textbf{Dynamic Focus}\\ 
\midrule
MSR-VTT~\cite{MSRVTT} & 1,000 & 15.01s & 9.41 & Human & \xmark & \xmark & \xmark\\
DiDeMo~\cite{DiDeMo} & 1,037 & 53.94s & 29.11 & Human & \xmark & \xmark & \xmark\\
MSVD~\cite{MSVD} & 670 & 10.04s & 7.01 & Human & \xmark & \xmark & \xmark\\
ActivityNet~\cite{ActivityNet} & 5,044 & 36.00s & 13.48 & Human & \xmark & \xmark & \xmark\\
DREAM-1K~\cite{tarsier} & 1,000 & 8.9s & 59.3 & Human & \xmark & \xmark & \cmark\\
VDC~\cite{AuroraCap} & 1,000 & 28.18s & 500.91 & GPT & \cmark & \cmark & \xmark\\
\midrule
\textbf{\benchmark} & 1,000 & 14.35s & 227.95 & Human & \cmark & \cmark & \cmark\\
\bottomrule
\end{tabular}
}
\caption{Comparison on statistics of retrieval and captioning benchmarks. All the statistics are reported on test split. Traditional benchmarks, namely MSR-VTT~\cite{MSRVTT}, MSVD~\cite{MSVD}, DiDeMo~\cite{DiDeMo} and ActivityNet~\cite{ActivityNet} have much shorter captions compared to \benchmark{}. Some detailed captioning benchmarks~\cite{tarsier, AuroraCap} have longer and detailed captions, but they are either annotated by GPT or do not focus on both static objects and dynamic actions. }
\label{table:benchmark-comparison}
\end{table*}

Video captioning~\cite{Git, mplug-2, tarsier, AuroraCap} and video retrieval~\cite{CLIP, CLIP4Clip, X-CLIP, VISTA, InternVideo2, LanguageBind, Long-CLIP} are two main tasks in video-language understanding. Video captioning requires models to perceive and describe the main objects, events and actions in the video, while retrieval aims at finding the most relevant video or text based on the text or video query. These two tasks can intuitively reflect the alignment degree and comprehension ability of Video-Language Models (VLMs) regarding videos and language, becoming the most crucial tasks for evaluating the capabilities of VLMs.

However, existing retrieval and captioning benchmarks have limitations in evaluating VLMs' fine-grained understanding level. Traditional benchmarks~\cite{MSRVTT, MSVD, DiDeMo} for retrieval and captioning have short and rough video descriptions annotated by human. These benchmarks effectively assess general video understanding in VLMs but fall short in evaluating fine-grained ability due to brief descriptions. Recently, some research (e.g., \cite{Long-CLIP, vript, AuroraCap}) makes use of powerful VLMs like GPT-4o~\cite{ChatGPT} to automate video annotation, which inevitably introduces the hallucinations and biases inherent in VLMs themselves. DREAM-1K~\cite{tarsier} adopts manual annotation to achieve a more accurate evaluation, yet it lacks diverse annotations.

In addition to the quality of annotations, designing effective metrics for video captioning also poses a challenge. Traditional metrics such as CIDEr~\cite{CIDEr} are difficult to be applied in the evaluation of fine-grained descriptions. Automated evaluation methods that utilize LLMs, such as AutoDQ \cite{tarsier} and VDCScore \cite{AuroraCap}, lack comprehensive consideration of both static objects and dynamic actions.

To address these problems, we present \benchmark{}, a fine-grained \textbf{Bench}mark designed for video \textbf{Ca}ptioning and \textbf{Re}trieval. It contains 1,000 videos with human-annotated detailed captions. Unlike image, video understanding tasks require models not only to understand the static scenes but also to grasp dynamic actions. With this in mind, we apply a hierarchical description scheme to the benchmark annotations. Each annotation covers four aspects: an overall summary, static object descriptions, dynamic action descriptions, and misc descriptions including filming style, camera movement, etc. Such a design ensures that each caption contains sufficient details, thereby challenging models to capture fine-grained information. Furthermore, to evaluate models spatiotemporally, each caption of \benchmark{} is manually separated into spatial parts and temporal parts. Based on this, we construct ReBias and CapST, two novel metrics tailored for the video retrieval and captioning tasks, respectively. These metrics give us a comprehensive insight into the spatiotemporal biases inherent in VLMs.

During the evaluation of powerful models on both video retrieval and video captioning tasks, we realize that previous research efforts treat video retrieval and video captioning as separate tasks, leading to the development of specialized models for each. Specifically, CLIP-based dual-encoder models have been advanced for video retrieval, while Multimodal Large Language Models (MLLMs) have been tailored for video captioning. However, we discover that video retrieval and video captioning can be unified and formulated as a mapping from the pixel space to a high-dimensional space: $\phi: \mathbb{R}^{T \times H \times W \times C} \rightarrow \mathbb{R}^D$ (either vocabulary space $\mathbb{R}^{D_v}$ or embedding space $\mathbb{R}^{D_e}$). This finding renders it feasible to address the gap between video retrieval and video captioning.

\begin{figure*}[htbp]
  \centering
  \includegraphics[width=1.05\linewidth]{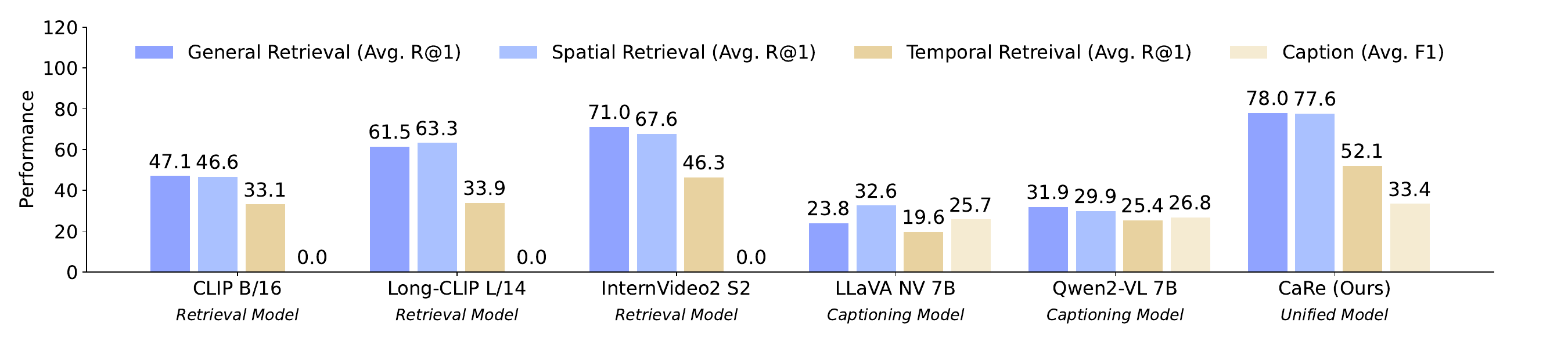}

   \caption{\textbf{Comparison on the \benchmark{} performance of CLIP-based retrieval models, MLLM captioning models and our unifed model.} The results on MLLMs are reported on their public version without contrastive training. The CLIP-based retrieval model has achieved excellent performance in video retrieval tasks, but it lacks the ability to describe videos. On the other hand, MLLM models are capable of describing videos in detail, but their retrieval performance is very poor. In contrast, CaRe, the unified model we propose, not only delivers outstanding performance in retrieval tasks but also has a strong capability to describe videos. Features are extracted from MLLMs using EOL prompt~\cite{E5-V}.}
   \label{fig:performance}
\end{figure*}

Taking advantage of the unified architecture of MLLMs, we develop \model{}, a simple and unified baseline capable of both detailed video captioning and fine-grained video retrieval. Specifically, our method involves a two-stage supervised fine-tuning (SFT). It equips the MLLM backbone with the unified ability of generating video captions and discriminating video contents. The first stage focuses on aligning the model output to a fine-grained text space, by training the model using mixed LLaVA-Video-178k~\cite{llava-video-178k} and Tarsier~\cite{tarsier} recaptioned data. In the second stage, a text-only contrastive learning approach~\cite{E5-V} is adopted to enable the MLLM to perform cross-modal representations. As shown in Figure \ref{fig:performance}, our experimental results indicate that, compared to CLIP-based retrieval models and MLLM captioning models, \model{} achieves superior performance on video captioning and retrieval tasks of \benchmark{}. 

In summary, we make the following contributions:

\begin{itemize}
\item We introduce a fine-grained testing benchmark named \benchmark{}. It is designed for video retrieval and video captioning, comprising 1,000 videos with high-quality human-annotated descriptions that provide sufficient video details. Each caption has four different aspects, ensuring that enough details are included. Uniquely, our \benchmark~provides manually separated spatial and temporal captions for each video, enabling us to independently test the spatiotemporal bias of VLMs. It challenges models to have an in-depth understanding of video contents. Based on this design, we construct ReBias and CapST, two novel metrics designed for the video retrieval and captioning tasks, respectively. 

\item We present \model, a simple and unified baseline for fine-grained video retrieval and captioning. By applying two-stage Supervised Fine-Tuning (SFT), we enable \model{} to not only generate detailed video descriptions but also to extract video features. Our experiment results show that, compared to the CLIP-based models designed for retrieval and the popular MLLMs skilled in video captioning, our baseline has competitive performance in both fine-grained video retrieval and detailed video captioning. 
\end{itemize}

\section{Related Work}
\label{sec:related}
\paragraph{Video Caption.} Video captioning aims to describe videos using natural language. It is a foundational task in video understanding. Early studies~\cite{Git, mplug-2} pretrain a VLM and finetune it on video captioning datasets with n-gram evaluation metrics such as CIDEr~\cite{CIDEr}. Traditional captioning benchmarks, such as ActivityNet~\cite{ActivityNet}, MSVD~\cite{MSVD}, and MSR-VTT~\cite{MSRVTT}, typically use a single sentence to describe the general content of a video clip. Consequently, the average caption length in these datasets is relatively short, making them insufficient to convey the full visual contents of videos. As a result, these traditional datasets can no longer effectively stress-test modern MLLMs, as these models are capable of generating captions that are more fine-grained than the existing ground truth.

To address these issues, new benchmarks have been proposed. For instance, DREAM-1K~\cite{tarsier} manually annotates five categories of videos rich in actions and introduces a novel automatic evaluation method, AutoDQ, to assess the accuracy and recall of actions and events in captions. Similarly, VDC~\cite{AuroraCap} employs a hierarchical prompting strategy to leverage GPT-4o in generating structured and detailed captions, followed by manual correction. It further evaluates caption accuracy along five dimensions, yet it does not explicitly consider human actions and motion. In this paper, we will explore a new fine-grained video captioning benchmark focusing not only on static objects but also dynamic actions, making it possible to comprehensively evaluate VLM's captioning performance.

\paragraph{Video Retrieval.}
Video retrieval aims at finding the most relevant video or text based on the text or video query. Traditional methods~\cite{InternVideo2, X-CLIP, CLIP4Clip, Long-CLIP, UMT, ImageBind} focus on using dual-encoder models based on CLIP~\cite{CLIP} to extract features of videos and texts. But most of these methods are limited by the 77-token context length inherited from CLIP and evaluated with short-caption benchmarks such as MSR-VTT~\cite{MSRVTT} and MSVD~\cite{MSVD}, making models difficult to understand long captions~\cite{VISTA}. As the field progresses, long-text and fine-grained video retrieval becomes important. Long-CLIP~\cite{Long-CLIP} is the first to address this problem. It trains CLIP on a context length of 248 to enable CLIP to handle long captions. But the benchmark used by Long-CLIP~\cite{Long-CLIP} are annotated by LLMs, which may contain coarse-grained, uncertain and wrong descriptions. In this paper, we will further explore the model training and the benchmark design in the fine-grained video retrieval task.

\paragraph{Multimodal Large Language Model.}
Due to the great advancements in LLMs~\cite{BERT, BiLLM, FLM_are_ZSL, PaLM}, their multimodal counterparts (MLLMs)~\cite{VideoChat, InternVL, MiniCPM-V, LLaVA-NeXT, Qwen2-VL}, are receiving significant attention, particularly for their capability to perform various visual tasks using straightforward instructions. Recent works like VideoChat~\cite{VideoChat} demonstrate outstanding performance on multimodal benchmarks such as Video-MME~\cite{Video-MME} and MVBench~\cite{MVBench}. But these models are restricted to generating responses based solely on user instructions and lack the capability to represent videos, images, and text. In this paper, we employ Qwen2-VL~\cite{Qwen2-VL} to construct a unified baseline that can handle both video retrieval and video captioning.

\paragraph{Multimodal Embedding.}
CLIP~\cite{CLIP} learns image and text representations by aligning them with contrastive learning. However, Mind the Gap~\cite{Mind-the-Gap} points out that different data modalities are embedded with gaps in their shared representation space. To address this issue, recent works like VISTA~\cite{VISTA} and E5-V~\cite{E5-V} begin to explore the possibilities of unified representation. They find that MLLMs provide a unified multimodal framework and can unify cross-modal representations without gaps. We regard it as a promising method and will explore further about unified MLLM representation on video retrieval.

\begin{figure}[!t]
  \centering
  \begin{subfigure}{0.48\linewidth}
    \includegraphics[width=\linewidth]{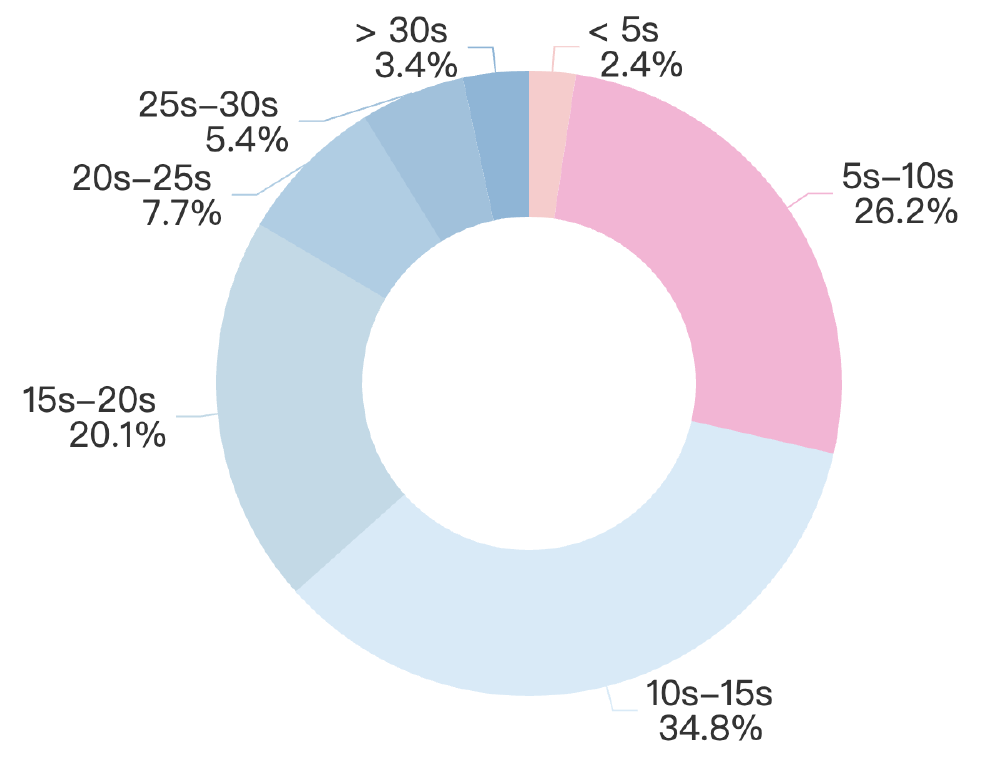}
    \caption{Video length distribution.}
    \label{fig:stats-video-length}
  \end{subfigure}
  \ \ \
  \begin{subfigure}{0.48\linewidth}
    \includegraphics[width=\linewidth]{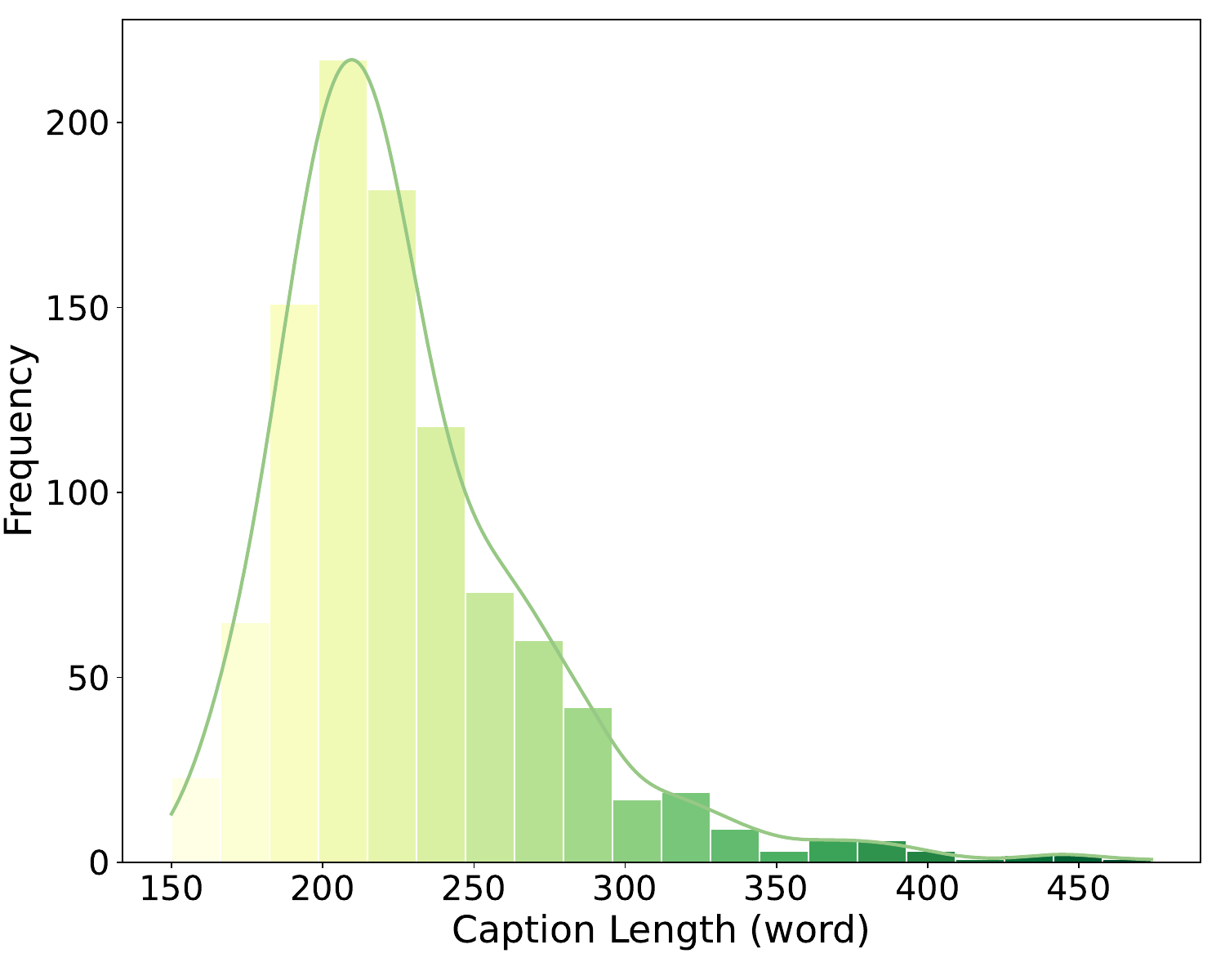}
    \caption{Caption length distribution.}
    \label{fig:stats-word-length}
  \end{subfigure}
  \caption{\textbf{Statistics of \benchmark{}.} Most videos range from 5-20 seconds and most captions fall between 150 and 300 words in length.}
  \label{fig:stats}
\end{figure}

\section{\benchmark{}: A Fine-Grained Benchmark}
\label{sec:benchmark}

\subsection{Video Collection}
\label{sec:collection}

We collect all videos from FineAction~\cite{FineAction}, a video dataset for temporal action localization with 106 subcategories and 4 major categories: \textit{personal care}, \textit{socializing \& relaxing}, \textit{sports \& exercise}, and \textit{household activities}. Videos in each subcategory share similar scenes and actions, which poses a challenge to the models' ability to understand and discriminate similar videos. 

We manually select 1,000 videos from FineAction~\cite{FineAction} with 10-20 videos in each subcategory. Videos are filtered out that \textbf{(1)} are not clear enough, \textbf{(2)} contain little actions and movements, and \textbf{(3)} contain vastly different scenes and actions which are easy for VLMs to discriminate.

\subsection{Two-Stage Annotation Pipeline}
\label{sec:pipeline}

\begin{figure*}[htbp]
  \centering
  \includegraphics[width=\linewidth]{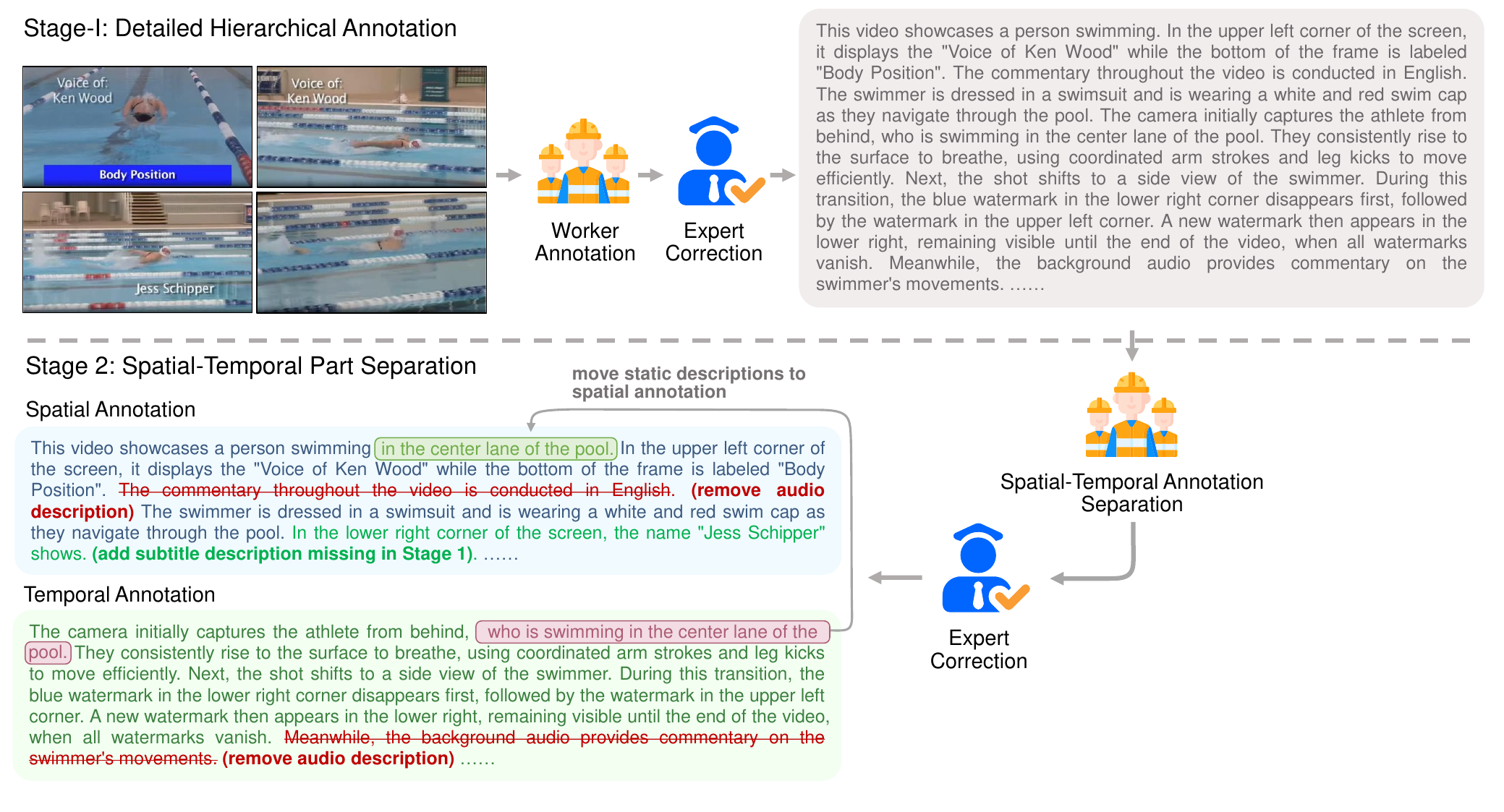}

   \caption{\textbf{An overview of the annotation pipeline.} In Stage-I, workers are asked to describe videos hierarchically in detail. In Stage-II, workers need to separate spatial descriptions with temporal descriptions. }
   \label{fig:pipeline}
\end{figure*}

The annotation pipeline consists of two stages. In the first stage, annotators are asked to generate detailed captions covering four key aspects of each video. Subsequently, they are guided to separate the annotations into temporal and spatial descriptions. To ensure high quality and minimize bias, each video is independently captioned by two annotators. Our experts subsequently refine and merge the captions after each stage. The annotation pipeline is illustrated in Figure \ref{fig:pipeline}.

\subsubsection{Stage-I: Detailed Annotation}

In Stage-I, annotators are tasked with describing videos in detail, with each description limited to 150-300 words to ensure conciseness and thoroughness. Each video description can be divided into four parts: a general overview, a spatial description, a object description, and an action description, as outlined below:

\begin{itemize}
    \item \textbf{General Overview} provides a one-sentence summary of the entire video. For example, \textit{this video shows a person slicing a watermelon.}
    \item \textbf{Object Description} focuses on static objects with attributes like position, color, shape, and other visual details. It contains primary and secondary objects, the background, their relative positions, interactions, and even visible elements such as watermarks.
    \item \textbf{Action Description} captures the actions occurring in the video, detailing the sequence of events (e.g., \textit{first..., then...}) and providing specific details of each action (e.g., \textit{rotating the watermelon clockwise}). It also includes the style of the actions (e.g., \textit{cutting fruit very quickly}, \textit{climbing the tree clumsily}).
    \item \textbf{Misc Description} is about 2-4 sentences in length. It covers different aspects, such as the viewpoint (e.g., \textit{This segment is filmed from a third-person perspective}) and the overall type of the video (e.g., \textit{providing a delightful and relaxing experience for viewers}).
\end{itemize}

\subsubsection{Stage-II: Spatio-Temporal Separation}
Stage-II refines the initial annotations by separating spatial and temporal elements. It removes action-related text (e.g., \textit{jump into the pool}) from object descriptions to create pure spatial descriptions, and eliminates static references (e.g., \textit{in the center lane of the pool}) from action descriptions to form pure temporal descriptions. This separation ensures precise evaluation of VLMs' spatial and temporal modeling capabilities by preventing interference between dynamic and static elements. 

\begin{itemize}
    \item \textbf{Spatial Description} provides a comprehensive view, beginning with a general overview and then detailing main objects, secondary objects, and the background environment. It ensures that spatial descriptions can differentiate between similar videos within the same subcategory.
    \item \textbf{Temporal Description} begins with a general overview, then focuses on actions and their order. Spatial-specific details are excluded. It ensures temporal descriptions uniquely identify each video within its subcategory.
\end{itemize}

Following the two stages, experts meticulously review and refine the results to ensure: \textbf{(1)} spatial and temporal annotations remain free of mixed action/object descriptions, \textbf{(2)} temporal descriptions include camera movements and subtitle changes, \textbf{(3)} subjective descriptions (e.g., \textit{the child looks very cute}) is eliminated, and \textbf{(4)} audio and speech references are excluded.

\subsection{Comparison on Statistics}

The captions in \benchmark{} are human-annotated, providing detailed and comprehensive descriptions of the videos. Consequently, its statistics differ significantly from those of traditional benchmarks. As shown in Table \ref{table:benchmark-comparison}, our benchmark is similar in size to MSR-VTT~\cite{MSRVTT}, DiDeMo~\cite{DiDeMo}, but the average number of words per caption is 24.2$\times$ higher than that of MSR-VTT~\cite{MSRVTT}, 7.82$\times$ higher than DiDeMo~\cite{DiDeMo}, and 32.5$\times$ higher than MSVD~\cite{MSVD}. The chart in Figure \ref{fig:stats-video-length} shows the video length distribution of \benchmark{}. Since excessively long video durations significantly increase the difficulty for annotators to provide detailed descriptions, our benchmark focuses on videos ranging from 5 to 20 seconds in length, with over 80\% of the videos falling within this range. Only 5.8\% are shorter than 5s or extends beyond 30s. Figure \ref{fig:stats-word-length} demonstrates how the caption length distributes. Most captions in \benchmark{} contain between 175 and 275 words.

\subsection{Metrics Design}

\benchmark{} contains manually annotated temporal and spatial captions. This design enables us to identify biases in the model's understanding of static objects and dynamic actions by analyzing the imbalance in spatiotemporal performance across video retrieval and captioning tasks. To quantify the spatiotemporal perfomance and bias, we introduce two novel metrics tailored for video retrieval and video captioning, respectively: ReBias and CapST. These two metrics allow us to comprehensively understand the VLMs' performance and inherent biases by separately benchmarking them on spatial tasks and temporal tasks.

\subsubsection{ReBias}

Evaluating spatial and temporal captions separately reveals the model's retrieval performance across both dimensions. By quantifying the imbalance in spatiotemporal retrieval performance, we can identify the model's bias towards its focus on static objects versus dynamic actions. Consequently, we introduce ReBias, a metric tailored to measure spatiotemporal \textbf{Re}trieval \textbf{Bias}. The formula for calculating this score is as follows:

\begin{equation}
    B = \left|1 - \frac{\Bar{R}_{temporal}}{\Bar{R}_{spatial}}\right|,
\end{equation}
where $\Bar{R}_{temporal}$ and $\Bar{R}_{spatial}$ denotes the average recall of R@1, R@5, R@10 on temporal and spatial retrieval, respectively.

ReBias measures a model's spatiotemporal imbalance by assessing how far the temporal-to-spatial recall ratio deviates from 1, effectively capturing its skew towards either dimension.

\subsubsection{CapST}
Traditional n-gram captioning metrics~\cite{CIDEr} fails to evaluate long and detailed captions. To overcome this limitation, we propose CapST, a video \textbf{Cap}tioning metric that comprehensively considers both static objects (\textbf{S}patial elements) and dynamic events (\textbf{T}emporal elements). Similar to \cite{tarsier}, a powerful LLM serves as an element extractor to extract events from temporal captions and objects from spatial captions. By computing the Natural Language Inference (NLI) relationship between the ground truth $D_{gt}$ and the predictions $D_{pred}$, we evaluate the quality of model-predicted descriptions. Specifically, we compute the recall and precision score: 
\begin{equation}
    R = \frac{N(D_{gt} \xrightarrow{\text{entail}} E_{pred})}{N(D_{gt})},
\end{equation}

\begin{equation}
    P = \frac{N(D_{pred} \xrightarrow{\text{entail}} E_{gt})}{N(D_{gt})},
\end{equation}
where $E_{pred}$ denotes elements (either objects or events) extracted from predictions, $E_{gt}$ denotes elements (either objects or events) extracted from ground truth captions, $N(D_{gt})$ is the number of ground truth captions, $N(D_{gt} \xrightarrow{\text{entail}} E_{pred})$ refers to the number of $E_{pred}$ entailed by $D_{gt}$, and $N(D_{pred} \xrightarrow{\text{entail}} E_{gt})$ means the number of $E_{gt}$ entailed by $D_{pred}$.

Specially, some static objects have multiple attributes, such as ``\textit{an elderly man wearing glasses and a blue suit.}" If the extracted object attributes are numerous and verbose, NLI may penalize predictions for not fully describing all attributes. To address this issue, we instruct the LLM to split attributes during extraction. For instance, the aforementioned description would be divided into ``\textit{an elderly man wearing glasses}" and ``\textit{an elderly man wearing a blue suit.}" This design allows a more precise evaluation of the model's performance to describe objects and their multiple attributes.

\begin{figure*}[!ht]
  \centering
  \includegraphics[width=\linewidth]{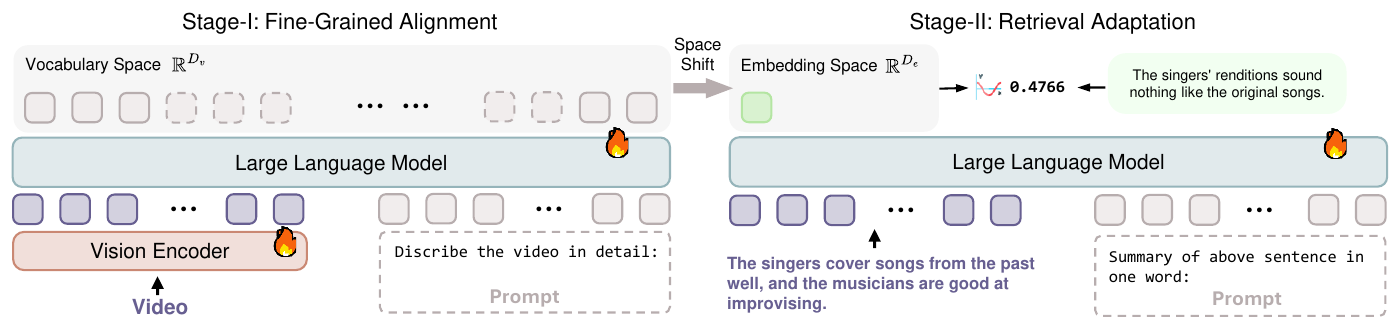}

   \caption{\textbf{The training recipe of \model{}.} In the first stage, we align \model{} outputs to a fine-grained text space, enabling it to describe videos in detail. In the second stage, a contrastive learning method is applied to get features from the inputs. The output space of \model{} shifts from the vocabulary space $\mathbb{R}^{D_v}$ in Stage-I to the embedding space $\mathbb{R}^{D_e}$ in Stage-II.}
   \label{fig:method}\
\end{figure*}

\section{\model{}: A Unified Video Model}
\label{sec:method}

Previous works treat video retrieval and captioning as separate tasks, fostering specialized models like CLIP-based dual-encoders for retrieval and MLLMs for captioning. However, we find that these tasks can be unified into a single framework, formulated as a mapping from the pixel space to a high-dimensional space: $\phi: \mathbb{R}^{T \times H \times W \times C} \rightarrow \mathbb{R}^D$ (either vocabulary space $\mathbb{R}^{D_v}$ or embedding space $\mathbb{R}^{D_e}$). To bridge this gap, we introduce \model{}, a unified baseline built on Qwen2-VL~\cite{Qwen2-VL}, trained via a two-stage progressive SFT to achieve both robust video captioning and strong video representation. The training pipeline is shown in Figure \ref{fig:method}.

\subsection{Stage-I: Fine-Grained Alignment} 
\label{sec:stage1}
MLLMs excel in generalization but often miss key video details or generate hallucinations. To align the model with fine-grained video understanding and provide a robust backbone for Stage-II, we train \model{} with high-quality video-caption pairs. Specifically, we set finetuning prompt to ``\texttt{Describe the video in detail.}" and train our model using some of video-text pairs from Tarsier Recap~\cite{tarsier}, emphasizing action-rich descriptions, and LLaVA-Video-178k~\cite{llava-video-178k}, focusing on short videos with detailed backgrounds. With fine-grained alignment, the model output is aligned with fine-grained text space and can focus on detailed actions and objects when describing videos.

\subsection{Stage-II: Retrieval Adaptation}
\label{sec:stage2}

After Stage-I training, \model{} achieves precise alignment between the pixel space and the fine-grained text space. To shift the model output from the vocabulary space $\mathbb{R}^{D_v}$ to the embedding space $\mathbb{R}^{D_e}$, we use a similar method as \cite{E5-V, PromptEOL}, employing an Explicit One-word Limitation (EOL) prompt to extract embeddings from \model{}. Specifically, there are two steps: (1) given an EOL prompt: ``\texttt{<sent> Summary of the above sentence in one word:}", the model is instructed to summarize the sentence $s_i$ in the next token; (2) we use the hidden states in the next token generation step as the final embeddings $f_i$. Then, we train the model on an NLI dataset~\cite{SimCSE} where each sample contains a sentence $s_i$, its positive $s_i^+$ and its hard negative $s_i^-$. Since there are no video inputs during Stage-II, we freeze the vision encoder and train the LLM only. Our training objective is given as:
\begin{equation}
\mathcal{L}=-\log \frac{e^{\cos \left(f_{i}, f_{i}^{+}\right) / \tau}}{\sum_{j=1}^{N}\left(e^{\cos \left(f_{i}, f_{j}^{+}\right) / \tau}+e^{\cos \left(f_{i}, f_{j}^{-}\right) / \tau}\right)},
\end{equation}
where $f_i$, $f_i^{+}$, $f_{i}^{-}$ denote the embeddings of the sentence $s_i$, its positive $s_i^+$ and its hard negative $s_i^-$, respectively. cos(·) is the cosine similarity function. $\tau$ is the temperature hyperparameter.

\section{Experiments}

\begin{table*}[t]
    \centering
    \setlength{\tabcolsep}{5pt}
    \resizebox{\textwidth}{!}{
    \begin{tabular}{lc|cccccccccc}
        \toprule

\multirow{3}{*}{\textbf{Model}} & \multirow{3}{*}{\textbf{\# Params}} & \multicolumn{10}{c}{\textbf{\benchmark{} Caption}} \\ \cmidrule{3-12}
\multicolumn{1}{c}{}  & \multicolumn{1}{c}{}  & \multicolumn{2}{|c}{\textbf{Personal Care}} & \multicolumn{2}{c}{\textbf{Socializing \& Relaxing}} & \multicolumn{2}{c}{\textbf{Sports \& Excercise}} & \multicolumn{2}{c}{\textbf{Household Activities}} & \multicolumn{2}{c}{\textbf{Overall}} \\
\multicolumn{1}{c}{}  & \multicolumn{1}{c}{}  & \multicolumn{1}{|c}{Action} & \multicolumn{1}{c}{Object} & \multicolumn{1}{c}{Action} & \multicolumn{1}{c}{Object} & \multicolumn{1}{c}{Action} & \multicolumn{1}{c}{Object} & \multicolumn{1}{c}{Action} & \multicolumn{1}{c}{Object} & \multicolumn{1}{c}{Action} & \multicolumn{1}{c}{Object} \\
\midrule
GPT-4o mini                       & -  &32.9/24.9/48.4 & 29.2/21.2/47.2 & 34.7/26.2/51.1 & 34.2/26.5/48.0 & 44.3/38.0/53.0 & 36.0/27.4/52.6 & 34.2/26.9/46.8 & 35.1/27.6/48.2& 36.8/29.1/50.2             & 33.8/25.8/49.1  \\
\midrule
LLaVA NV~\cite{LLaVA-NeXT}     & 7B  &27.5/20.1/43.7 & 21.7/15.5/36.2 & 25.0/17.4/44.1 & 24.1/17.3/39.9 & 29.4/21.1/48.4 & 26.8/19.6/42.3 & 24.3/16.2/\underline{48.1} & 26.3/19.5/40.4& 26.6/18.7/45.9             & 24.7/17.9/39.8  \\
InternVL2~\cite{InternVL}      & 7B  &22.2/18.4/28.0 & 20.4/15.1/31.6 & 23.0/17.9/32.3 & 23.1/17.3/34.6 & 27.9/23.4/34.5 & 24.9/18.3/38.7 & 18.4/14.7/24.8 & 22.7/17.1/33.8& 23.3/18.8/30.7             & 22.9/17.1/34.9  \\
InternVL2.5~\cite{InternVL2.5}                       & 7B  &22.0/15.1/41.1 & 26.4/20.4/37.2 & 24.0/16.8/41.6 & 28.4/\underline{22.7}/37.9 & 34.0/26.1/48.8 & 31.6/\underline{26.4}/39.4 & 22.3/15.3/40.6 & 29.6/\underline{24.4}/37.7& 26.0/18.6/43.2             & 29.1/\underline{23.5}/38.2  \\
InternVL2.5~\cite{InternVL2.5}                        & 72B &24.6/16.7/46.7 & 28.7/\underline{22.4}/40.0 & 25.9/18.3/44.4 & 28.6/\textbf{23.3}/37.3 & 36.0/27.8/51.0 & \textbf{34.0}/\textbf{28.2}/42.7 & 24.9/17.5/43.2 & 30.8/\textbf{25.7}/38.5& 28.2/20.3/46.4             & 30.5/\textbf{24.8}/39.5  \\
MiniCPM-V 2.6~\cite{MiniCPM-V}    & 7B  &30.2/21.3/52.0 & 28.9/19.7/53.6 & 26.9/18.6/48.8 & 29.4/21.0/48.8 & 38.1/29.7/53.1 & 32.0/23.7/49.3 & 28.5/20.0/\textbf{49.5} & 32.2/23.3/52.1 & 31.1/22.3/51.2             & 30.5/21.9/50.5  \\
Tarsier~\cite{tarsier}                           & 7B  & 25.4/16.5/\textbf{55.0} & 30.0/22.2/45.9 & 26.5/18.0/\textbf{50.4} & 30.0/22.6/44.4 & 32.0/22.8/53.3 & 33.4/24.9/50.7 & 22.8/15.3/44.7 & 31.2/23.9/45.1 & 27.1/18.4/51.1 & 31.1/23.4/46.5  \\
Qwen2-VL~\cite{Qwen2-VL}                          & 7B  &28.4/23.9/34.9 & 23.7/15.8/47.7 & 27.5/20.8/40.3 & 23.0/15.1/47.8 & 33.0/26.6/43.6 & 24.9/16.2/53.1 & 25.7/20.2/35.1 & 24.8/16.8/47.2 & 28.8/22.9/39.0             & 24.0/15.9/49.1  \\
Qwen2-VL~\cite{Qwen2-VL}                             & 72B &29.6/22.1/45.0 & 24.5/16.3/49.4 & 28.1/20.6/44.2 & 22.5/14.7/47.8 & 37.3/28.5/53.9 & 24.6/15.8/56.3 & 26.4/18.6/45.4 & 26.5/17.4/55.7 & 30.5/22.6/47.1             & 24.2/15.8 /51.9  \\
\midrule
\textbf{\model$_{\mathrm{stage-I}}$\;}     & 7B  & \underline{33.9}/\underline{25.4}/50.8 & \textbf{32.1}/\textbf{22.6}/\underline{55.3} & \textbf{32.4}/\textbf{24.0}/\underline{49.8} & \underline{31.3}/22.2/\underline{53.1} & \textbf{42.8}/\textbf{33.7}/\textbf{58.5} & \underline{33.2}/23.2/\underline{58.4} & \textbf{31.5}/\textbf{24.4}/44.7 & \textbf{33.6}/23.8/\textbf{57.1} & \textbf{35.3}/\textbf{26.9}/\underline{51.3} & \textbf{32.4}/22.9/\underline{55.7}  \\
\textbf{\model\;}                       & 7B  & \textbf{34.4}/\textbf{25.6}/\underline{52.6} & \underline{30.9}/21.1/\textbf{57.2} & \underline{32.2}/\textbf{24.0}/48.8 & \textbf{31.5}/21.9/\textbf{55.6} & \underline{42.3}/\underline{33.3}/\underline{58.1} & 31.8/21.3/\textbf{62.6} & \underline{30.9}/\underline{23.4}/45.3 & \underline{32.6}/23.0/\underline{55.8}& 35.1/\underline{26.6}/\textbf{51.4}    & \underline{31.7}/21.8/\textbf{57.8}  \\
\bottomrule
\end{tabular}

}
\caption{\textbf{Video caption performance of popular state-of-the-art models on \benchmark{}.} We report F1/Recall/Precision for each category. LLaVA NV is short for LLaVA NeXT Video. \# Params denotes the number of LLM parameters. Deepseek-V3\cite{DeepSeekV3} serves as the LLM judge.} 
\label{table:reca_captioning_leaderboard}
\vspace{5mm}
\end{table*}

\begin{table}[t]
\centering
\setlength{\tabcolsep}{5pt}
\resizebox{0.5\textwidth}{!}{
\begin{tabular}{l|cccccc}
\toprule

\multirow{3}{*}{\textbf{Model}} & \multicolumn{6}{c}{\textbf{\benchmark{} General Retrieval}} \\ \cmidrule{2-7}
\multicolumn{1}{c}{}  & \multicolumn{3}{|c}{Text-to-Video} & \multicolumn{3}{c}{Video-to-Text} \\
\multicolumn{1}{c}{} & \multicolumn{1}{|c}{\textbf{R@1}}      & \textbf{R@5}     & \textbf{R@10}   & \textbf{R@1}& \textbf{R@5}& \textbf{R@10}\\

\midrule

\multicolumn{7}{c}{\textbf{CLIP-based Models}} \\ \midrule
CLIP B/16~\cite{CLIP}                            & 45.7          & 79.6          & 89.1          & 48.4          & 82.4          & 90.8   \\
CLIP L/14~\cite{CLIP}                            & 51.2          & 83.4          & 90.6          & 54.7          & 86.9          & 93.6   \\
LanguageBind~\cite{LanguageBind}                 & 64.3          & 91.0          & 96.3          & 59.5          & 88.0          & 95.0   \\
Long-CLIP B/14~\cite{Long-CLIP}                  & 59.2          & 85.3          & 92.1          & 55.8          & 84.7          & 92.9   \\
Long-CLIP L/14~\cite{Long-CLIP}                  & 62.7          & 88.8          & 95.7          & 60.3          & 88.8          & 94.9   \\
InternVideo2$_{stage2}$ 1B~\cite{InternVideo2}   & 72.5          & 93.7          & 97.3          & 69.5          & 94.6          & 97.8   \\

\midrule
\multicolumn{7}{c}{\textbf{MLLMs}} \\
\midrule
LLaVA NV 7B~\cite{LLaVA-NeXT}                    & 22.4          & 51.5          & 65.3           & 25.2         & 54.4          & 67.7 \\
MiniCPM-V 2.6~\cite{MiniCPM-V}                   & 8.2           & 26.9          & 38.4           & 16.7         & 39.9          & 55.8 \\
InternVL2 8B~\cite{InternVL}                     & 34.6          & 67.1          & 80.2           & 35.1         & 68.5          & 82.0 \\
Tarsier 7B~\cite{tarsier}                        & 26.8          & 64.6          & 83.5           & 32.3         & 68.0          & 84.4 \\
Qwen2-VL 7B~\cite{Qwen2-VL}                      & 30.9          & 64.7          & 79.1           & 32.9         & 69.6          & 82.7 \\
\midrule
\multicolumn{7}{c}{\textbf{Contrastively trained MLLMs}} \\
\midrule
LLaVA NV 7B~\cite{LLaVA-NeXT}                    & 66.9          & 89.4          & 96.0          & 62.7          & 89.2          & 95.4 \\
MiniCPM-V 2.6~\cite{MiniCPM-V}                   & 71.0          & 92.2          & 97.0          & 69.3          & 92.8          & 97.1 \\
InternVL2 8B~\cite{InternVL}                     & 72.1          & 92.6          & 96.8          & 73.6          & 93.4          & 97.4   \\
Tarsier 7B~\cite{tarsier}                        & 71.0          & 93.8          & 97.8          & 70.6          & 94.2          & 98.0  \\
Qwen2-VL 7B~\cite{Qwen2-VL}                      & 76.6          & 95.3          & \textbf{98.7} & 77.4          & 95.6          & 98.7  \\
\midrule
\textbf{\model}                                  & \textbf{77.0} & \textbf{95.6} & \textbf{98.7} & \textbf{79.0} & \textbf{96.8} & \textbf{99.1} \\
\bottomrule

\end{tabular}
}
\caption{\textbf{Video retrieval performance of some state-of-the-arts methods on \benchmark{}.} LLaVA NV is short for LLaVA NeXT Video. We train all the MLLMs contrastively on NLI dataset to enable them to generate video embeddings. All the results are reported in zero-shot setting.} 
\label{table:retrieval_leaderboard}
\end{table}

\begin{table*}[t]
\begin{threeparttable}
\centering
\resizebox{\textwidth}{!}{
\begin{tabular}{l|cccccccccccc|c}
\toprule
\multirow{3}{*}{\textbf{Model}} & \multicolumn{6}{c}{\textbf{\benchmark{} Spatial Retrieval}} & \multicolumn{6}{c|}{\textbf{\benchmark{} Temporal Retrieval}} & \multirow{3}{*}{\textbf{ReBias}\%$\downarrow$} \\ \cmidrule{2-13}
\multicolumn{1}{c}{}  & \multicolumn{3}{|c}{Text-to-Video} & \multicolumn{3}{c}{Video-to-Text} & \multicolumn{3}{c}{Text-to-Video} &\multicolumn{3}{c|}{Video-to-Text} & \multicolumn{1}{c}{} \\
\multicolumn{1}{c|}{} & \textbf{R@1} &\textbf{R@5} & \textbf{R@10} &\textbf{R@1} &\textbf{R@5} & \textbf{R@10} & \textbf{R@1}   & \textbf{R@5}     & \textbf{R@10}  & \textbf{R@1}       & \textbf{R@5} & \textbf{R@10} & \multicolumn{1}{c}{}\\
\midrule

\multicolumn{14}{c}{\textbf{CLIP-based Models}} \\ \midrule
CLIP B/16~\cite{CLIP}                                            & 45.6 & 79.0 & 89.2 & 47.6 & 80.9 & 90.8 & 30.3 & 65.1 & 79.8 & 35.8 & 71.0 & 85.8 & 17.75 \\
CLIP L/14~\cite{CLIP}                                            & 49.0 & 81.9 & 91.4 & 55.4 & 85.6 & 93.0 & 33.5 & 70.3 & 84.0 & 39.7 & 76.2 & 87.9 & 16.52 \\
LanguageBind~\cite{LanguageBind}                                 & 64.7 & 90.8 & 96.8 & 61.0 & 87.2 & 94.5 & 39.8 & 77.3 & 90.5 & 42.2 & 77.6 & 91.7 & 18.10 \\
Long-CLIP B/14~\cite{Long-CLIP}                                  & 62.5 & 86.0 & 92.7 & 53.8 & 84.1 & 92.7 & 32.0 & 65.4 & 79.3 & 29.7 & 67.3 & 84.1 & 31.88 \\
Long-CLIP L/14~\cite{Long-CLIP}                                  & 65.6 & 90.9 & 96.0 & 61.0 & 88.3 & 94.4 & 33.2 & 68.8 & 81.6 & 34.5 & 71.9 & 86.6 & 31.77 \\
$\text{InternVideo2}_{stage2}$ 1B~\cite{InternVideo2}$^\dagger$  & 72.4 & 94.2 & 97.4 & 62.7 & 90.5 & 95.9 & 46.0 & 80.8 & 91.9 & 46.6 & 82.5 & 92.5 & 16.58 \\

\midrule
\multicolumn{14}{c}{\textbf{MLLMs}} \\
\midrule
LLaVA NV 7B~\cite{LLaVA-NeXT}    & 34.1 & 63.1 & 76.0 & 31.1 & 63.7 & 75.1 & 18.6 & 48.1 & 62.4 & 20.7 & 47.1 & 62.4 & 32.32 \\
MiniCPM-V 2.6~\cite{MiniCPM-V}   & 6.6  & 25.2 & 35.7 & 13.3 & 38.2 & 53.5 & 11.8 & 35.8 & 52.2 & 16.6 & 47.4 & 64.4 & 24.41 \\
InternVL2 8B~\cite{InternVL}     & 40.4 & 72.9 & 83.8 & 40.3 & 73.0 & 85.7 & 29.3 & 62.5 & 77.4 & 27.1 & 59.8 & 75.9 & 19.31 \\
Tarsier 7B~\cite{tarsier}        & 40.5 & 74.0 & 88.1 & 41.9 & 75.0 & 87.4 & 26.8 & 64.6 & 83.5 & 32.3 & 68.0 & 84.4 & 13.15\\
Qwen2-VL 7B~\cite{Qwen2-VL}      & 28.1 & 61.3 & 76.1 & 31.6 & 65.6 & 80.4 & 24.3 & 61.5 & 78.4 & 26.4 & 59.2 & 76.1 & 5.28 \\
\midrule
\multicolumn{14}{c}{\textbf{Contrastively trained MLLMs}} \\
\midrule
LLaVA NV 7B~\cite{LLaVA-NeXT}    & 68.0 & 92.0 & 96.2 & 65.0 & 90.0 & 95.9 & 43.3 & 76.9 & 88.9 & 40.1 & 75.4 & 88.7  & 22.69 \\
MiniCPM-V 2.6~\cite{MiniCPM-V}   & 71.7 & 93.6 & 98.0 & 67.6 & 92.3 & 97.7 & 50.5 & 82.9 & 92.1 & 46.1 & 80.9 & 93.3 & 16.89 \\
InternVL2 8B~\cite{InternVL}     & 76.1 & 94.1 & 97.6 & 74.3 & 94.5 & 97.6 & 48.1 & 76.8 & 89.0 & 47.6 & 78.2 & 90.3 & 25.02 \\
Tarsier 7B~\cite{tarsier}        & 70.2 & 94.0 & 98.2 & 67.4 & 93.5 & 97.4 & 50.1 & 84.1 & 92.8 & 50.0 & 84.7 & 94.9 & 14.04 \\
Qwen2-VL 7B~\cite{Qwen2-VL}      & \textbf{78.2} & \underline{95.5} & \underline{98.5} & \underline{75.4} & \underline{95.0} & \underline{98.1} & \textbf{51.9} & \underline{84.8} & \textbf{94.9} & \underline{52.7} & \underline{85.4} & \textbf{95.2} & 16.30 \\
\midrule
\model{}                         & \underline{76.8} & \textbf{96.3} & \textbf{98.7} & \textbf{78.1} & \textbf{95.8} & \textbf{99.3} & \underline{50.7} & \textbf{85.3} & \underline{94.4} & \textbf{53.4} & \textbf{86.3} & \underline{94.0} & 17.53 \\
\bottomrule
\end{tabular}
}
\begin{tablenotes}
\small
\item[$\dagger$] $\text{InternVideo2}_{stage2}$ is tested without match header for fairness.
\end{tablenotes}
\end{threeparttable}
\caption{\textbf{Spatiotemporal retrieval results of video retrieval on \benchmark{}.} LLaVA NV 7B is short for LLaVA NeXT Video 7B. We train all the MLLMs contrastively on NLI dataset to enable them to generate video embeddings. All the results are reported in zero-shot setting.} 
\label{table:retrieval_spatiotemporal_leaderboard}
\end{table*}

\subsection{Settings}
\label{sec:settings}

Our experiments are conducted on 8 NVIDIA H800 80G (Stage-I) and 8 NVIDIA RTX A6000 48G (Stage-II). In Stage-I, we adapt the public Qwen2-VL~\cite{Qwen2-VL}, training it with a learning rate of 2e-5, a batch size of 64, a max pixel of 460,800, and 16 input frames. For Stage-II, \model{}$_{\textrm{stage-II}}$ is initialized from Stage-I and trained on the NLI dataset with the video backbone frozen. We set the epoch, batch size, and warmup ratio to 2, 768, and 0.2, respectively, and fully fine-tune \model{}$_{\textrm{stage-II}}$ with a learning rate of 2e-4.

\subsection{Video Captioning}

In Table \ref{table:reca_captioning_leaderboard}, we present quantitative comparison of the video captioning task on \benchmark{} between \model{} and existing state-of-the-art methods. All the results are reported in zero-shot setting following our CapST metric. We employ DeepSeek-V3~\cite{DeepSeekV3} to serve as the LLM judge, as it not only delivers precise judgment but also has lower costs compared to ChatGPT~\cite{ChatGPT}. For fairness, the number of input frames are set to 32. The default prompt is ``\texttt{Describe the video in detail.}" unless the official research~\cite{LLaVA-NeXT} recommends a specific one. 

As illustrated in Table \ref{table:reca_captioning_leaderboard}, our model has demonstrated superior performance across all the categories, surpassing all existing open-source models currently available. Considering the disparity between the models' parameters and their performance, even the most powerful MLLM, Qwen2-VL 72B, which stands as a pioneer in the realm of open-source models, exhibits a significant performance gap when compared to our 7B \model{}. This indicates that all current models have yet to achieve the capability of providing highly detailed, comprehensive, and fine-grained descriptions of videos. Additionally, it can be observed that whether the model has undergone stage II training does not affect its captioning performance. These promising results demonstrate that even a small-scale 7B model is capable of understanding the details within videos, including dynamic actions and static object elements and can have outstanding captioning and retrieval abilities at the same time.

\subsection{Video Retrieval}

We compare CLIP-based models, contrastively trained MLLMs and our \model{} on \benchmark{}, following the setting of 32 input frames. Table \ref{table:retrieval_leaderboard} and Table \ref{table:retrieval_spatiotemporal_leaderboard} present the general retrieval performance and spatiotemporal retrieval performance on \benchmark{}. General retrieval uses first-stage annotations, while spatial and temporal retrieval leverage spatial captions and temporal captions from second-stage. All tasks employ Recall at Rank K (R@K, higher is better) in a zero-shot setting. The following observations can be concluded according to our analysis:

\begin{enumerate}
    \item \textbf{MLLMs perform better than CLIP-based models on video retrieval.} CLIP-based models have long dominated retrieval performance benchmarks. However, as demonstrated in Table \ref{table:retrieval_leaderboard}, MLLMs trained with contrastive learning exhibit significantly enhanced retrieval capabilities, surpassing their predecessors in performance. Our \model{} yields the most favorable results, surpassing CLIP, Long-CLIP, LanguageBind, InternVideo2 and all the other MLLMs.
    \item \textbf{All models have inherent biases in their spatiotemporal understanding} According to spatiotemporal retrieval results in Table \ref{table:retrieval_spatiotemporal_leaderboard}, all models exhibit imbalance in spatiotemporal understanding, with spatial retrieval performance significantly outperforming temporal retrieval performance. This indicates that models are more inclined to comprehend static objects rather than distinguishing videos by focusing on the dynamic actions. Such a bias highlights the need for improved methodologies to enhance temporal understanding capabilities in video understanding tasks.
    
\end{enumerate}

\subsection{Ablation Study}

In this section, we conduct experiments to further investigate the effect of our proposed two-stage SFT. Using the same setting as mentioned in Section \ref{sec:settings} and building upon the Qwen2-VL model~\cite{Qwen2-VL}, we perform a quantitative analysis to evaluate the impact of Stage-I and Stage-II on the model's performance in video captioning and video retrieval tasks. The results are shown in Table \ref{table:ablation}. Our baseline model, Qwen2-VL~\cite{Qwen2-VL}, shows strong captioning skills \textcolor{gray}{(Avg. F1 26.8)} but struggles with retrieval tasks \textcolor{gray}{(Avg. R@1 25.6)} without retrieval adaptation training. Adding fine-grained alignment training greatly improves the model's captioning ability \textcolor{gray}{(Avg. F1 +7.0)} at a slight cost to retrieval performance \textcolor{gray}{(Avg. R@1 -8.0)}. On the other hand, just using retrieval rdaptation training gives the model excellent retrieval capabilities \textcolor{gray}{(Avg. R@1 +51.4)}, which is a big improvement over the baseline. After completing both training stages, our model not only performs well in detailed video description but also achieves top-level retrieval performance. Interestingly, we have uncovered evidence that video retrieval and video captioning tasks can mutually enhance each other: retrieval adaptation improves the baseline's video captioning performance by \textbf{+1.4} \textcolor{gray}{(Avg. F1 from 26.8 to 28.2)}, and the high-quality fine-tuning of fine-grained alignment further boosts the retrieval adapted model by \textbf{+1} \textcolor{gray}{(Avg. R@1 from 77.0 to 78.0)}.
\begin{table}
\centering
\resizebox{0.48\textwidth}{!}{
\begin{tabular}{llll}
\toprule
\multirow{2}{*}{\textbf{Setting}} & \textbf{Retrieval} & \textbf{Caption} & \textbf{Overall} \\
\multicolumn{1}{c}{}    & Avg. R@1 & Avg. F1 & Unified Score\\ 
\midrule
Baseline & 25.6 & 26.8 & 26.2\\
+Align & 17.6\textcolor{gray}{(-8.0)} & 33.8\textcolor{gray}{(+7.0)} & 25.7\textcolor{gray}{(-0.5)}\\
+Adaptation & 77.0\textcolor{gray}{(+51.4)} & 28.2\textcolor{gray}{(+1.4)} & 52.6\textcolor{gray}{(+26.4)}\\
\midrule
+Align \& Adaptation & 78.0\textcolor{gray}{(+52.4)} & 33.4\textcolor{gray}{(+6.6)} & 55.7\textcolor{gray}{(+29.5)} \\
\bottomrule
\end{tabular}
}
\caption{\textbf{Effect of the two-stage training.} Four model settings are included: the baseline, \model{} with only fine-grained alignment, \model{} with only retrieval adaptation, and \model{} trained with the full two-stage SFT. The evaluation metrics include Avg. R@1, which denotes the average text-to-video and video-to-text R@1 on \benchmark{} General Retrieval, and Avg. F1, which represents the average action and object F1 score on the \benchmark{} Captioning. The unified score is the average of R@1 and F1.} 
\label{table:ablation}
\end{table}

\subsection{Logits Visualization}

\begin{figure*}[htbp]
    \centering

    \begin{minipage}[t]{0.98\textwidth}
        \centering
        \includegraphics[width=\linewidth]{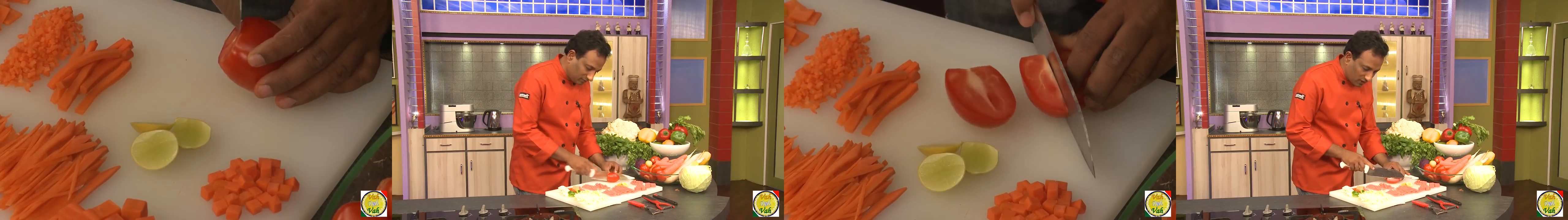}
        \subcaption{The input video.}
        \label{fig:logits_input_video}
    \end{minipage}
    
    \begin{minipage}[t]{\textwidth}
        \centering
        \includegraphics[width=\linewidth]{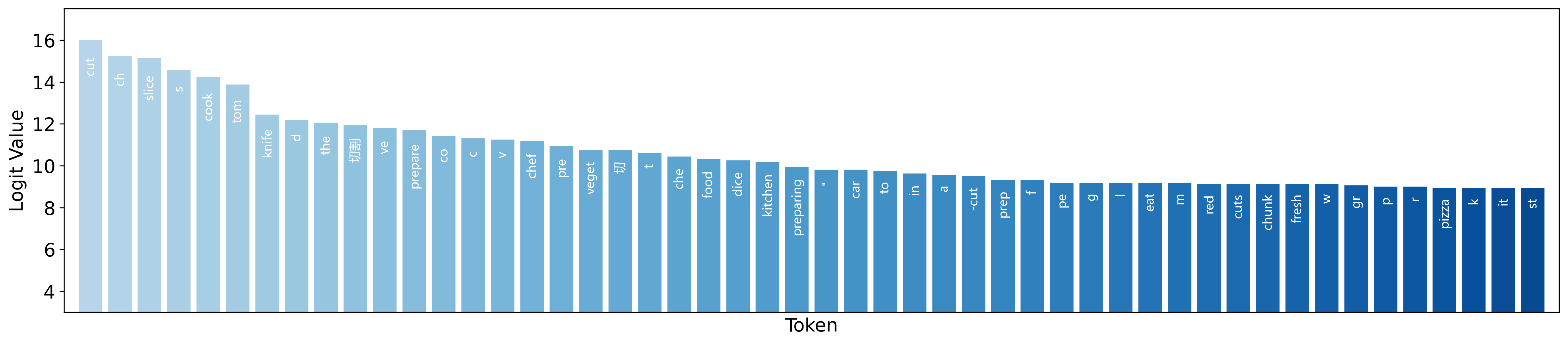}
        \subcaption{Top 50 tokens decoded from the output embeddings of Qwen2-VL.}
        \label{fig:logits_baseline}
    \end{minipage}
    
    \begin{minipage}[t]{\textwidth}
        \centering
        \includegraphics[width=\linewidth]{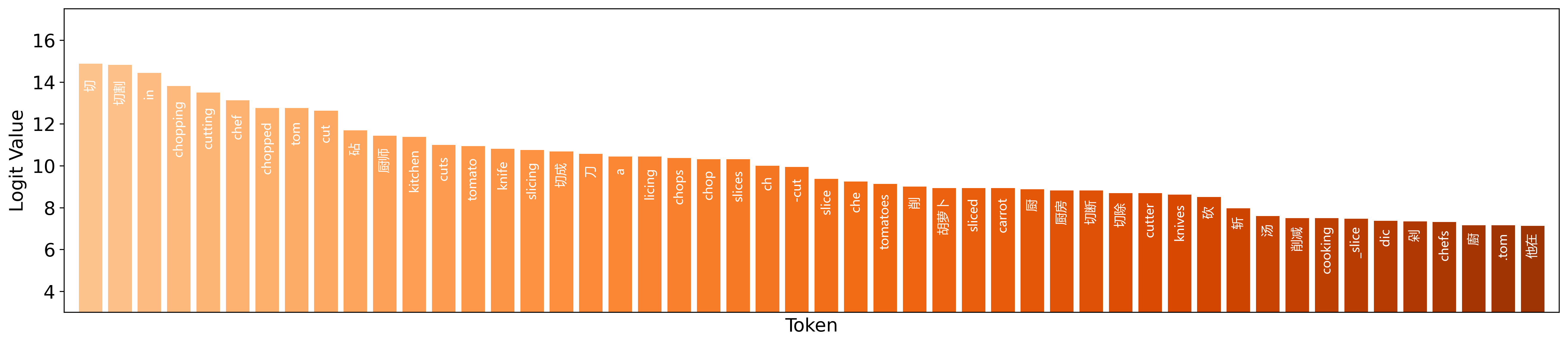}
        \subcaption{Top 50 tokens decoded from the output embeddings of \model{}.}
        \label{fig:logits_care}
    \end{minipage}
    
    \caption{\textbf{Top 50 tokens decoded from the output embeddings of Qwen2-VL and \model{}.} Qwen2-VL is the baseline model of \model{} without any SFT. Compared to Qwen2-VL, two-stage SFT makes the semantic components of \model{} embedding much more related to the input video featuring \textit{a chef is cutting tomatoes in the kitchen}.}
    \label{fig:logits_visualization}
\end{figure*}

To explore how \model{} works, we feed its output embedding of a video featuring \textit{a chef is cutting tomatoes in the kitchen} into the last linear layer (i.e. lm\_head). It projects the embedding into the vocabulary space. By decode the output logits, we can easily visualize the semantic components of an embedding. It can be discovered that tokens with high logits constitute the essential semantics of the input video, as shown in Figure \ref{fig:logits_care}, describing the main visual objects and actions of the video such as \textit{kitchen}, \textit{cutting}, \textit{tomatoes} and \textit{chef}, while the tokens in Figure \ref{fig:logits_baseline} contain many subwords and irrelevant tokens like \textit{dice}, \textit{car} and \textit{pizza}. It can be inferred that the semantic distribution in the next token space is hugely changed by two-stage SFT, allowing the main semantics to be the core components of the embedding.

\section{Conclusion}

In this work, we present \benchmark{}, a fine-grained benchmark for video captioning and retrieval, featuring 1,000 videos with high-quality human-annotated descriptions. Each caption is structured hierarchically to cover four key aspects: overall summary, static object descriptions, dynamic action descriptions, and miscellaneous details such as filming style and camera movement. We also propose ReBias and CapST, novel metrics for assessing retrieval and captioning performance. Additionally, we develop \model{}, a unified baseline for both tasks, leveraging a two-stage supervised fine-tuning approach to generate detailed captions and extract video features. Experiments show that \model{} outperforms specialized models in both fine-grained retrieval and captioning. Our work highlights the potential of unifying video captioning and retrieval tasks under a single framework, challenging the traditional methods. However, our model doesn't address problems about VLMs' bias towards the focus on static objects and dynamic actions. Look ahead, future research could explore further integration of both tasks and try to develop a more balanced model.

{
    \small
    \bibliographystyle{ieeenat_fullname}
    \bibliography{main}
}
\clearpage
\setcounter{page}{1}
\appendix

\begin{table*}[t]
\begin{threeparttable}
\centering

\resizebox{0.9\textwidth}{!}{
\begin{tabular}{l|cccccccccccc}
    \toprule

\multirow{3}{*}{Model} & \multicolumn{6}{c}{MSR-VTT~\cite{MSRVTT}} & \multicolumn{6}{c}{MSVD~\cite{MSVD}} \\ \cmidrule{2-13}
\multicolumn{1}{c}{} & \multicolumn{3}{|c}{Text-to-Video} & \multicolumn{3}{c}{Video-to-Text} & \multicolumn{3}{c}{Text-to-Video} &\multicolumn{3}{c}{Video-to-Text} \\
\multicolumn{1}{c|}{} & \textbf{R@1}       & \textbf{R@5}     & \textbf{R@10}   & \textbf{R@1}& \textbf{R@5}& \textbf{R@10}& \textbf{R@1} &\textbf{R@5} & \textbf{R@10} &\textbf{R@1} &\textbf{R@5} & \textbf{R@10}\\
\midrule
\multicolumn{13}{c}{\textbf{CLIP-based Models}} \\ \midrule
CLIP B/16~\cite{CLIP}             & 33.8        & 56.1     & 66.6     & 30.5  & 53.8  & 65.5 & 37.0 & 64.2	& 74.1 & 60.5 &	79.9 & 87.5\\
CLIP L/14~\cite{CLIP}             & 36.7        & 58.8     & 68.0  & 32.8  & 54.7  & 66.2 & 41.1 & 68.8 & 77.5 & 68.1 & 85.5 & 91.8\\
LanguageBind~\cite{LanguageBind}  & 42.1  & 65.9 & 75.5 & 40.1 & 65.4 & 73.9 & 50.0 & 77.7 & 85.6 & 75.1 & 90 & 94.2\\
Long-CLIP B/14~\cite{Long-CLIP}   & 38.7 & 62.3 & 70.6 & 34.4 & 57.7 & 68.2 & 40.4 & 68.0 & 77.7 & 63.4 & 81.6 & 87.8\\
Long-CLIP L/14~\cite{Long-CLIP}   & 40.9 & 65.5 & 74.6 & 36.2 & 62.2 & 71.5 & 46.5 & 73.5 & 82.0 & 69.3 & 86.0 & 90.3 \\
$\text{InternVideo2}_{stage2}$ 1B~\cite{InternVideo2}$^\dagger$  & 44.2 & 70.1 & 78.1 & 40.5 & 66.9 & 76.3 & 53.0 & 79.1 & 87.2 & 74.6 & 88.5 & 93.4\\

\midrule
\multicolumn{13}{c}{\textbf{Contrastively Trained MLLMs}} \\ \midrule
LLaVA NV 7B~\cite{LLaVA-NeXT}  & 40.3 & 64.9 & 74.1 &	30.5 & 58.0 & 69.0 & 47.3 & 75.7 & 83.7 & 51.9 & 74.3 & 81.8 \\
InternVL2 8B~\cite{InternVL}    & 44.6 & 69.3 & 77.4 & 40.8 & 66.6 & 76.5 & 47.7 & 75.9 & 83.9 & 64.2 & 81.3 & 87.2 \\
MiniCPM-V 2.6~\cite{MiniCPM-V}   & 44.7 & 69.7 & 77.8 & 41.6 & 68.7 & 77.6 & 50.5 & 78.7 & 85.8 & 69.1 & 84.6 & 90.2 \\
Tarsier 7B~\cite{tarsier}       & 43.4 & 69.2 & 77.0 & 35.8 & 62.5 & 72.3 & 52.1 & 79.7 & 86.5 & 67.8 & 88.8 & 93.1 \\
Qwen2-VL 7B~\cite{tarsier}       & 46.9 & 69.2 & 79.7 & 43.4 & 69.2 & 78.8 & 53.3 & 79.7 & 86.5 & 73.7 & 89.6 & 92.4 \\
\midrule
\model{}                         & 43.9 & 67.0 & 75.7 & 41.7 & 68.1 & 76.2 & 52.6 & 79.2 & 86.6 & 74.6 & 87.9 & 92.4 \\
\bottomrule
\end{tabular}
}
\begin{tablenotes}
\small
\item[$\dagger$] $\text{InternVideo2}_{stage2}$ is tested without match header for fairness.
\end{tablenotes}
\end{threeparttable}
\caption{\textbf{Results of video retrieval on MSR-VTT~\cite{MSRVTT} and MSVD~\cite{MSVD}.} LLaVA NV is short for LLaVA NeXT Video. All the results are reported in zero-shot setting.} 
\label{table:msrvtt_msvd_retrieval_leaderboard}
\end{table*}

\begin{table}[t]
\begin{threeparttable}
\centering
\setlength{\tabcolsep}{5pt}
\resizebox{0.5\textwidth}{!}{
\begin{tabular}{l|cccccc}
    \toprule

\multirow{3}{*}{Model} & \multicolumn{6}{c}{DiDeMo} \\ \cmidrule{2-7}
\multicolumn{1}{c}{}  & \multicolumn{3}{|c}{Text-to-Video} & \multicolumn{3}{c}{Video-to-Text} \\
\multicolumn{1}{c|}{} & \textbf{R@1}       & \textbf{R@5}     & \textbf{R@10}   & \textbf{R@1}& \textbf{R@5}& \textbf{R@10}\\
\midrule

\multicolumn{7}{c}{\textbf{CLIP-based Models}} \\ \midrule
CLIP B/16~\cite{CLIP}                                  & 23.5 & 46.3 & 55.2 & 22.2 & 43.8 & 54.0\\
CLIP L/14~\cite{CLIP}                                  & 24.1 & 48.0 & 58.2 & 23.8 & 44.9 & 54.0\\
LanguageBind~\cite{LanguageBind}                       & 35.6 & 63.6 & 71.7 & 35.6 & 62.8 & 71.8\\
Long-CLIP B/14~\cite{Long-CLIP}                        & 30.3 & 52.4 & 63.7 & 24.8 & 52.8 & 63.4\\
Long-CLIP L/14~\cite{Long-CLIP}                        & 32.4 & 56.2 & 65.2 & 28.5 & 54.1 & 64.7\\
$\text{InternVideo2}_{stage2}$ 1B~\cite{InternVideo2}$^\dagger$  & 35.0 & 63.7 & 74.1 & 35.5 & 60.7 & 70.7\\

\midrule
\multicolumn{7}{c}{\textbf{Contrastively Trained MLLMs}} \\ \midrule
LLaVA NV 7B~\cite{LLaVA-NeXT}    & 36.0             & 62.3             & 71.7             & 31.4             & 58.0             & 68.0            \\
InternVL2 8B~\cite{InternVL}     & 39.7             & 65.6             & 74.1             & 35.5             & 64.0             & 72.2            \\
MiniCPM-V 2.6~\cite{MiniCPM-V}   & 40.6             & 65.2             & 74.2             & 35.7             & 61.6             & 70.1            \\
Tarsier 7B~\cite{tarsier}        & 42.1             & 68.2             & 77.1             & 39.5             & 64.6             & 73.7            \\
Qwen2-VL 7B~\cite{tarsier}       & 46.1             & 69.6             & 77.6             & 42.1             & 66.1             & 76.3            \\
\midrule
\model{}                         & 41.4             & 68.5             & 77.1             & 39.1             & 66.0             & 75.8            \\
\bottomrule
\end{tabular}
}
\begin{tablenotes}
\small
\item[$\dagger$] $\text{InternVideo2}_{stage2}$ is tested without match header for fairness.
\end{tablenotes}
\end{threeparttable}

\caption{\textbf{Results of video retrieval on DiDeMo~\cite{DiDeMo}.} LLaVA NV is short for LLaVA NeXT Video. All the results are
reported in zero-shot setting.} 
\label{table:didemo_retrieval_leaderboard}
\end{table}

\section{Additional Experiments}

We compare CLIP-based models, MLLMs, and \model{} on traditional retrieval benchmarks. All the experiments follow the setting of 32 input frames. Table \ref{table:msrvtt_msvd_retrieval_leaderboard} and Table \ref{table:didemo_retrieval_leaderboard} present the retrieval performance of all the models on MSR-VTT~\cite{MSRVTT}, MSVD~\cite{MSVD} and DiDeMo~\cite{DiDeMo}. All the results are reported in zero-shot setting.

\section{Annotation Guideline}
\label{sec:guideline}

To inform our annotators the key points that they need to pay attention to, we design a guideline to teach them how to describe videos accurately. The guideline is shown below.

\begin{textbox}{Annotation Guideline (Stage 1)}

\textbf{\ul{Task}}

Your task is to describe videos in detail and hierarchically within 150-300 words. We provide two examples and some points you may need to know.

\textbf{\ul{Example 1: Cutting a Watermelon}}

\textit{(A video about cutting a watermelon is provided.)}

\begin{itemize}
    \item \textbf{Summary}\; This video shows a man cutting a watermelon.
    \item \textbf{Object Description}\; The man is wearing a green T-shirt and a black apron, with a black mesh hat on his head. His left hand is wearing a gray glove, while his right hand, holding a fruit knife, is wearing a transparent glove. He stands at the corner of the countertop, with a white cutting board in front of him, holding a watermelon. To his left, there is a sink containing another uncut watermelon.
    \item \textbf{Action Description}\; The man first cuts off both ends of the watermelon. Then, he places the watermelon \ul{\textit{upright}} and \ul{\textit{rotate it clockwise}}, slicing off the rind piece by piece. He uses the knife to push the rind into a trash bin \ul{\textit{on his right}}. Next, he takes a light green tray from his right and place it next to the cutting board. After peeling the watermelon, he cuts it into pieces and slides them onto the light green tray.
    \item \textbf{Misc Description}\; The video is filmed from behind the man, showing a quick and efficient process of cutting the watermelon. With impressive speed, he slices through the fruit, showing his expertise.
\end{itemize}

\textbf{\ul{Example 2: Cutting a Tomato}}

\textit{(A video about cutting a tomato is provided.)}

\begin{itemize}
    \item \textbf{Summary}\; In the footage, someone is holding a knife and cutting a tomato on a cutting board.
    \item \textbf{Object Description}\; The person is wearing black clothes, with a watch on his left wrist. On the cutting board, there are four previously cut tomatoes and one sliced green fruit. On the table, there is a bag of uncut tomatoes and a small knife. \ul{\textit{In the top left corner of the video, there is a "luxeat" watermark, and the text “NOW I'VE SEEN EVERYTHING” is written in the bottom left corner.}}
    \item \textbf{Action Description}\; While cutting the tomato, the person first slices it forcefully with one cut, then \textit{speeds up the chopping frequency}, quickly slicing the tomato into neat pieces.
    \item \textbf{Misc Description}\; The video is filmed from a third-person perspective, showcasing clean and efficient vegetable-cutting. The person’s motions are skillful and confident.
\end{itemize}
\textbf{\ul{Key Points for Descriptions}}
\begin{itemize}
    \item \textbf{Object Description}\; Describe the entire frame in as much detail as possible. Focus on the objects visible in the frame, clearly describing their positions, appearances, and interactions (e.g., ``left hand" ``right hand" ``on the left" ``on the right" ``above" ``below" ``upside-down" ``holding" ``wearing" etc.). This part should follow the description order outlined below: (1) describe the main object in the frames: for example, ``The person is wearing a green T-shirt and a black apron, with a black mesh hat on their head. His left hand is wearing a gray glove, while his right hand, holding a fruit knife, is wearing a transparent glove." (2) describe the secondary objects in the frames: for example, ``The person is standing at one corner of a metal countertop. In front of him is a white cutting board with a watermelon on it. To his left, there is a sink containing another uncut watermelon."
    \item \textbf{Action Description}\; Clearly describe the actions performed by the main subject, noting the sequence of events (e.g., first do X, then do Y). Include details about the nuances of the actions (e.g., rotating the watermelon clockwise, flipping it upside-down) and the style of execution (e.g., cutting fruit very quickly, climbing a tree clumsily). 
    \item \textbf{Misc Description} \;Describe the video’s filming perspective (e.g., “first-person,” “third-person,” “off-site footage of a competition”) and provide a brief summary of the overall style and impression conveyed by the actions (e.g., orderly and fast watermelon cutting, sharp and efficient movements, clumsy actions, or dangerous behaviors). This part should be concise, within 2-4 sentences.
\end{itemize}
\end{textbox}

\begin{textbox}{Annotation Guideline (Stage 2)}

\textbf{\ul{Task}}

In this stage, your task is to separate the original hierarchical descriptions into two parts: spatial descriptions (which do not include any descriptions about movements) and temporal descriptions (which do not include any object descriptions). Camera movements, such as zoom-ins, zoom-outs, etc., should be included in temporal descriptions.

\textbf{\ul{Key Points for Descriptions}}

\begin{itemize}
    \item The spatial description should cover the key objects, secondary objects, and the environment in the frame. It must ensure that, based on the spatial description alone, the videos in the assigned subcategory can be differentiated from one another.
    \item The temporal description should exclude any obvious static object descriptions that help distinguish different videos. Only the details and sequence of actions should be kept, and it must ensure that, based on the temporal description alone, the videos in the assigned subcategory can be differentiated from one another.
    \item All the contents of spatial and temporal descriptions should come from the Stage 1 descriptions, and no additional details should be added. Both spatial and temporal descriptions should begin with a summary.
\end{itemize}

\end{textbox}

\section{Case study}

Benchmarks like MSRVTT~\cite{MSRVTT} rely on brief short captions. As shown in Figure \ref{fig:caption-compare}, the MSRVTT caption in the upper-left corner overlooks key details, such as the contents of the kitchen and the attire of the man. Captions annotated by LLMs may have coarse-grained, uncertain and wrong descriptions. As shown in Figure \ref{fig:caption-compare}, GPT-4o erroneously identifies the slipper beneath the phone as a phone case and describes the camera's violent shaking as ``minimal movement." The fine-grained caption on the right is selected from \benchmark{} and is created by human. The \textcolor[RGB]{101,165,66}{green} sentences are fine-grained descriptions and the \textcolor[RGB]{179,86,40}{brown} words show the action sequences in the video. For more sample of \benchmark{}, see the end of this supplementary materials.

\clearpage

\begin{figure*}[t]
  \centering
  \includegraphics[width=0.9\linewidth]{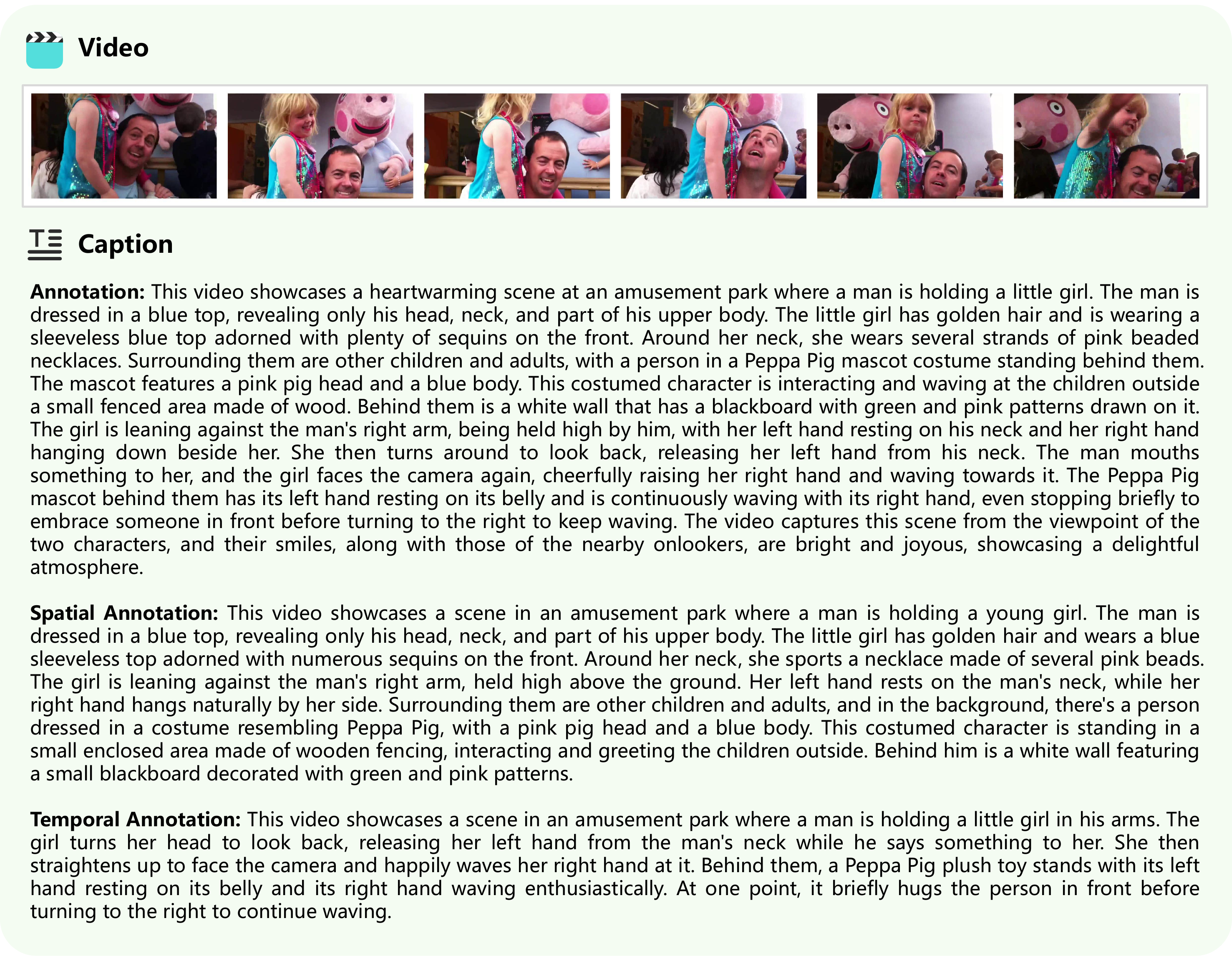}
  \includegraphics[width=0.9\linewidth]{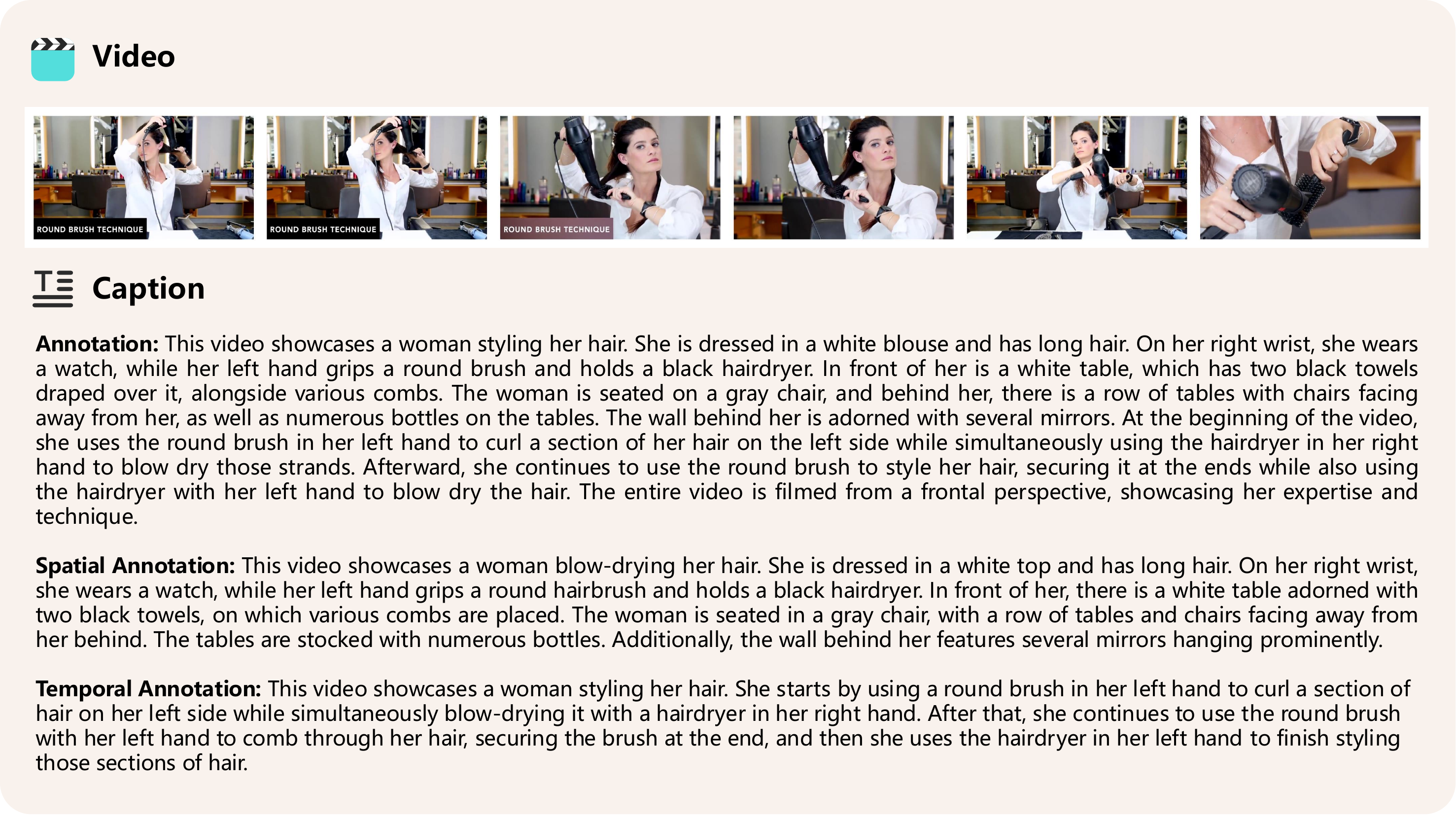}
\end{figure*}

\begin{figure*}[t]
  \centering
  \vspace{-5mm}
  \includegraphics[width=0.9\linewidth]{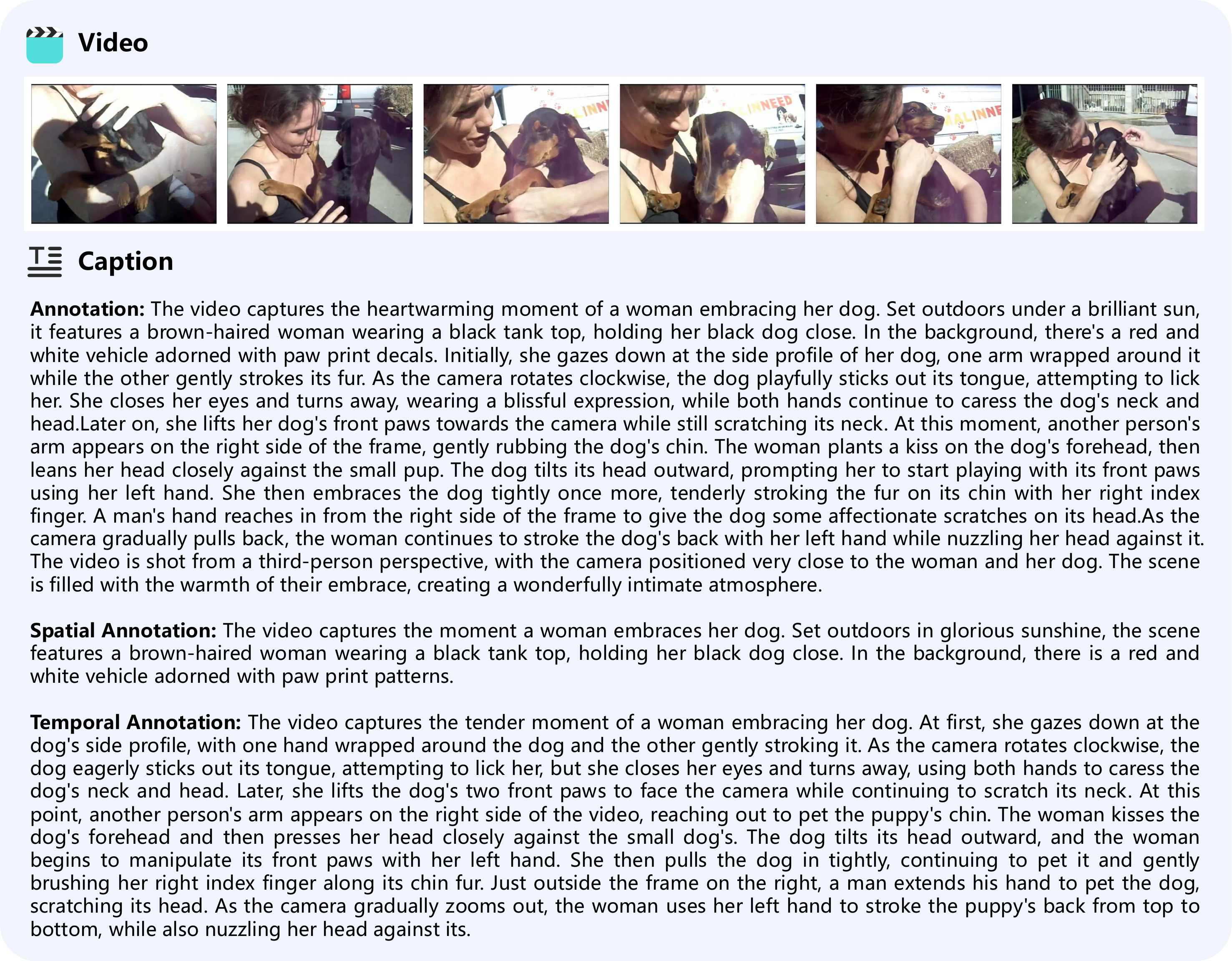}
  \includegraphics[width=0.9\linewidth]{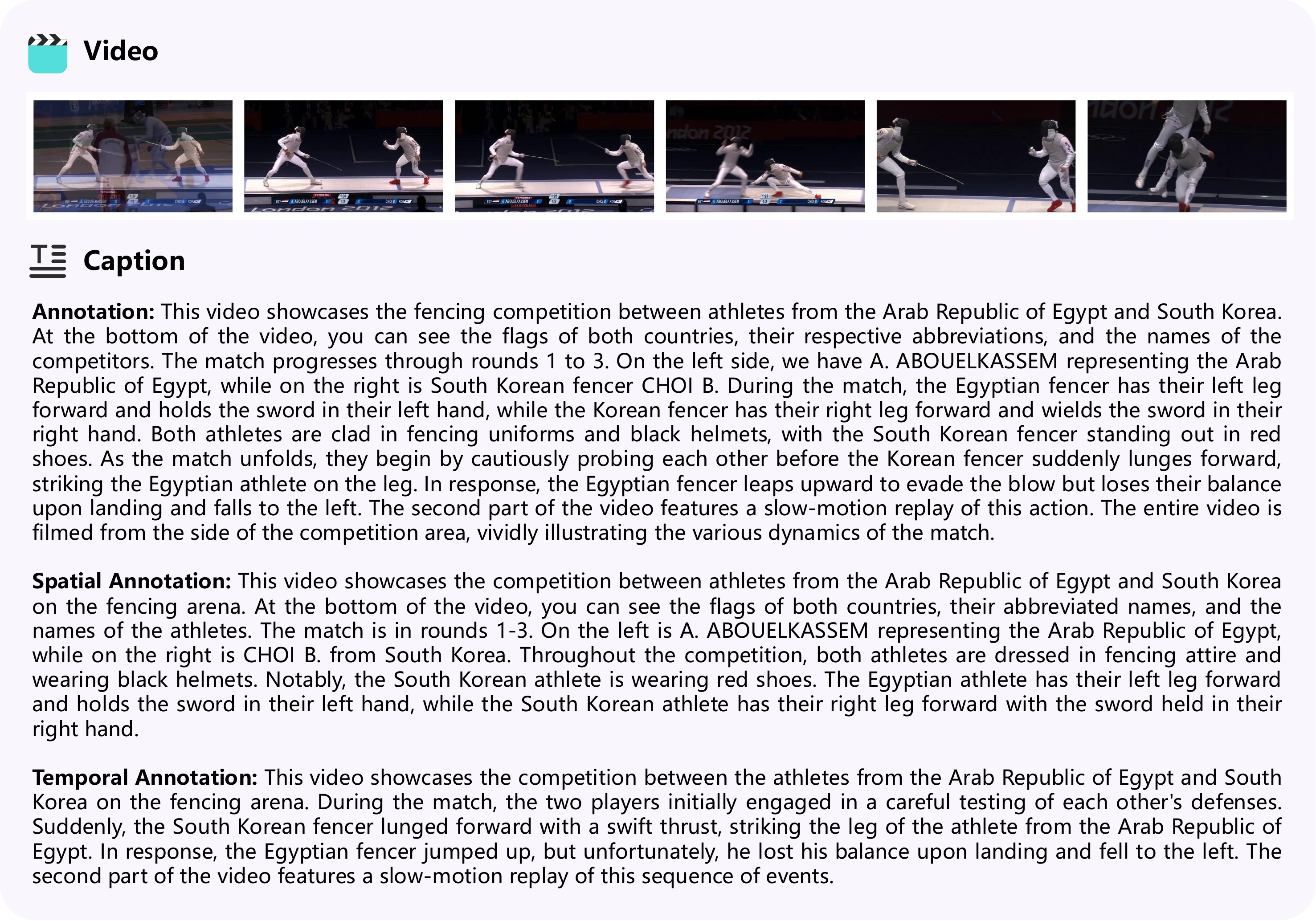}
\end{figure*}

\end{document}